\documentclass[runningheads]{llncs}

\usepackage{eccv}

\usepackage{eccvabbrv}

\usepackage{multirow}
\usepackage{graphicx}
\usepackage{booktabs}
\usepackage{hyperref}
\usepackage{booktabs}
\usepackage{orcidlink}
\usepackage{colortbl}  
\usepackage{array}   
\usepackage{xcolor}
\usepackage[accsupp]{axessibility}  

\usepackage{hyperref}
\hypersetup{breaklinks=true,colorlinks,citecolor=black}

\usepackage{orcidlink}

\begin{document}

\title{SCAPE: A Simple and Strong Category-Agnostic Pose Estimator\thanks{Yujia Liang 
 and Zixuan Ye contributed equally. Hao Lu is the corresponding author.}} 


\author{Yujia Liang\orcidlink{0009-0009-1432-3461} \and
Zixuan Ye\orcidlink{0000-0001-8517-682X} \and
Wenze Liu\orcidlink{0000-0002-1510-6196} \and
Hao Lu\orcidlink{0000-0003-3854-8664}}

\authorrunning{Y. Liang et al.}

\institute{
National Key Laboratory of Multispectral Information Intelligent Processing Technology; School of Artificial Intelligence and Automation, Huazhong University of Science and Technology, China
\\
\email{\{yjl,hlu\}@hust.edu.cn}}

\maketitle
\begin{abstract}

Category-Agnostic Pose Estimation (CAPE) aims to localize keypoints on an object of any category given few exemplars in an in-context manner. Prior arts involve sophisticated designs, \textit{e.g.}, sundry modules for similarity calculation and a two-stage framework, or takes in extra heatmap generation and supervision. We notice that CAPE is essentially a task about feature matching, which can be solved within the attention process. Therefore we first streamline the architecture into a simple baseline consisting of several pure self-attention layers and an MLP regression head---this simplification means that one only needs to consider
the attention quality to boost the performance of CAPE. Towards an effective attention process for CAPE, we further introduce two key modules: i) a global keypoint feature perceptor to inject global semantic information into support keypoints, and ii) a keypoint attention refiner to enhance inter-node correlation between keypoints. They jointly form a Simple and strong Category-Agnostic Pose Estimator (SCAPE). Experimental results show that SCAPE outperforms prior arts by $2.2$ and $1.3$ PCK under $1$-shot and $5$-shot settings with faster inference speed and lighter model capacity, excelling in both accuracy and efficiency. Code and models are available at \href{https://github.com/tiny-smart/SCAPE}{github.com/tiny-smart/SCAPE}.

\end{abstract}

\section{Introduction}
\label{sec:intro}
\begin{figure}[!t]
\centering{\includegraphics[width=1.0\linewidth]{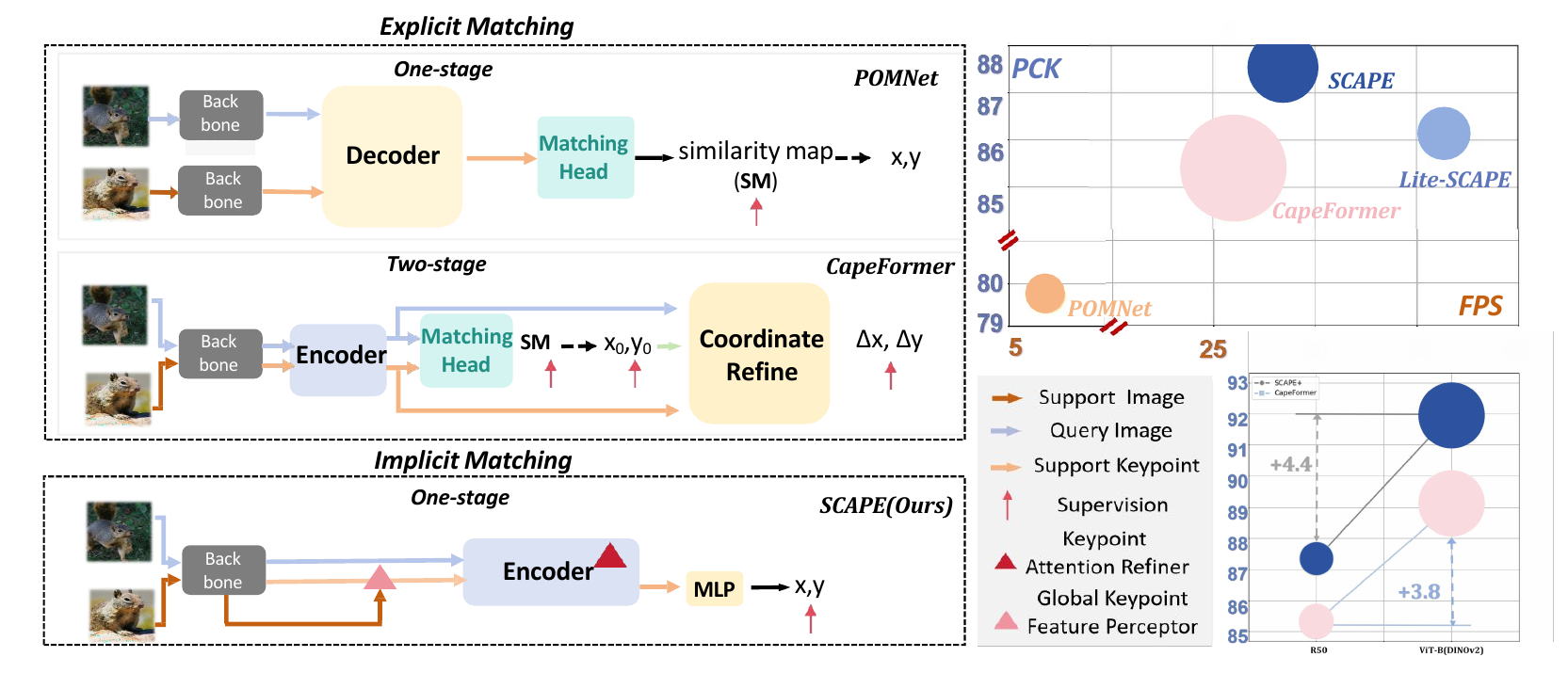}}
    \caption{\textbf{Comparison with prior arts.} (a) POMNet~\cite{xu2022pose} relies on similarity matching to obtain similarity maps and infer keypoint coordinates. (b) CapeFormer~\cite{shi2023matching} presents a two-stage framework, iteratively refining unreliable initial predictions. (c) Our SCAPE employs self-attention for feature interaction and directly regresses keypoints without explicit matching. For better similarity matching, we introduce two modules. The circle size indicates the model parameters (excluding backbone). The y-axis represents the accuracy (PCK), and the x-axis indicates the inference speed (FPS). Our model facilitates the seamless integration of state-of-the-art self-supervised learning techniques for scaling Vision Transformers (ViTs).} 
\label{fig:compare}
\end{figure}

\begin{figure}[!t]
\centering{\includegraphics[width=0.6\linewidth]{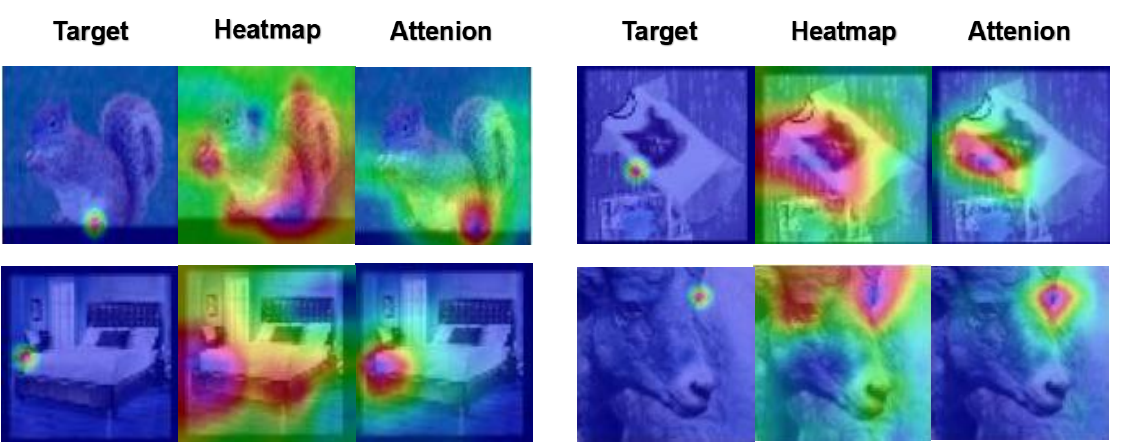}}
    \caption{\textbf{Implicit attention map is closer to ground truth than explicit similarity map.} The second column represents the similarity map obtained from CapeFormer, and the third column denotes the final layer of attention map between support keypoints and query image in the first stage of CapeFormer.}
\label{fig:cross_attention}
\end{figure}

Category-Agnostic Pose Estimation (CAPE), introduced by Xu~\etal~\cite{xu2022pose}, is a recent emerging topic in pose estimation. It extends conventional category-specific pose estimation~\cite{newell2016stacked,xiao2018simple,sun2019deep,labuguen2021macaquepose} and multi-category pose estimation~\cite{xu2022vitpose+} to unseen categories given few image-annotation examples. 
In their preliminary solution POMNet\cite{xu2022pose}, CAPE is regarded as a similarity matching problem, 
solved by generating a similarity map for keypoint prediction. Under a similar 
vein, CapeFormer~\cite{shi2023matching} adopts an additional transformer decoder~\cite{liu2022dab} to iteratively refine matched keypoints in a two-stage manner. Besides applying self-attention blocks to extract image features, their models further involve a series of 
sophisticated modules for feature matching. Their
pipelines 
are shown in Fig.~\ref{fig:compare}. Actually, there is a common sense 
that transformer attention is already a strong operator for similarity calculation and matching, suggesting redundancy in previous models and matching policies. Per Fig.~\ref{fig:cross_attention}, an initial exploration of CapeFormer~\cite{shi2023matching} shows that the attention maps exhibit even closer responses to ground truth before extra 
stage-two 
post-processing. 
This reveals that extra complex blocks or additional supervision 
may be redundant or suboptimal. 

\begin{figure}[!t]
\centering{\includegraphics[width=1.0\linewidth]{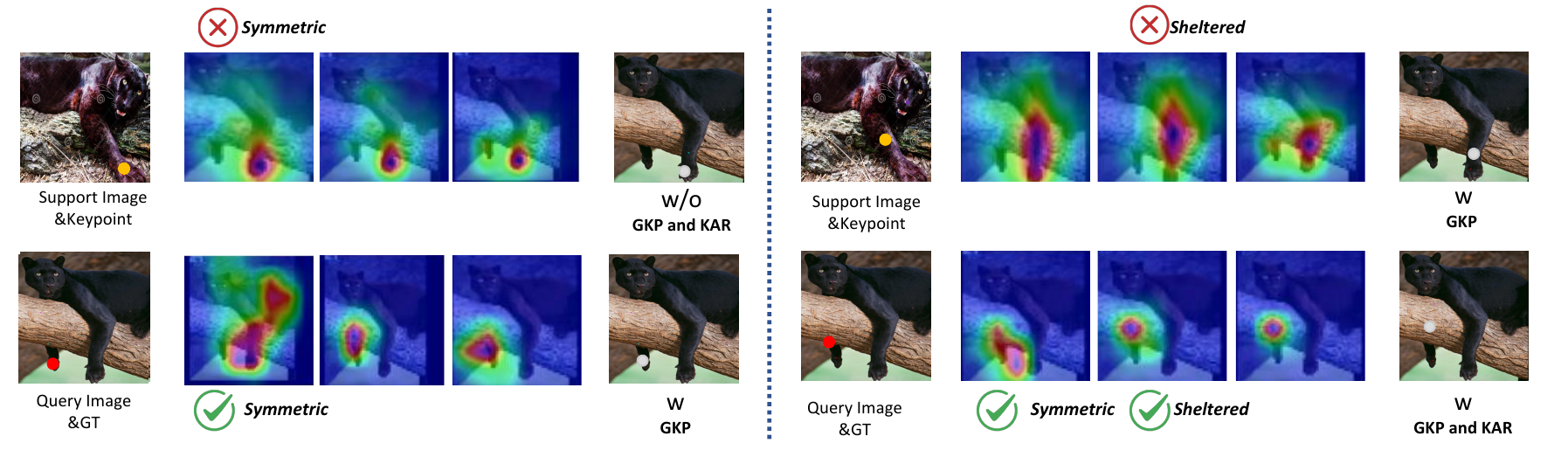}}
    \caption{\textbf{Visualization of the last three attention maps between keypoints and query image and result}. (left) The query target is the right foot; however, the attention easily focuses on the left foot with a similar appearance, leading to inaccurate estimation. GKP injects global information for the support keypoints, equipping it with relative positional information to distinguish left and right keypoints. (right) However, GKP struggles when facing shelter. By modeling the correlation between keypoints, we enable the inference from visible to invisible. Despite the right knee is sheltered, the model can infer its correct position by locating the right foot.}
\label{fig:6}

\end{figure}

We thus streamline the pipeline, and build a simple baseline including several self-attention layers and an MLP head, where the output coordinates are directly supervised. This simplification makes it easier to improve the localization accuracy of CAPE, by only 
concerning about the attention quality in self-attention. On this basis, we then share insights to fit the internal attention maps right for CAPE. We begin with two representative hard cases in CAPE: symmetric keypoints and sheltered keypoints. i) Symmetric keypoints, due to their similar appearance, are difficult to be distinguished by the model. As shown in Fig.~\ref{fig:6}, the attention mechanism struggles to capture correct keypoint positions and turns to other confusing keypoint choices. To enhance the discrimination ability for symmetric keypoints, we propose a Global Keypoint feature Perceptor (GKP) module to fuse the global information from the support image into support keypoint. By leveraging the surrounding information, the attention map can easily identify keypoints that have symmetric counterparts or similar neighbor keypoints. ii) Sheltered keypoints are another difficult case, 
where attention fails to find the invisible target keypoint. A natural solution is to infer its position with the assistance of other visible correlated keypoints. Therefore, we 
attempt to address this issue by establishing the intrinsic correlation among all keypoints, whose necessity and benefit have also been verified in category-specific pose estimation (CSPE)~\cite{ma2021context,li2021tokenpose,liu2023group}. We thus introduce the Keypoint Attention Refiner (KAR) module, which infuses the intrinsic relationship among keypoints as a refinement for the attention map. KAR effectively enables the reasoning power from visible to sheltered keypoints, as shown in Fig.~\ref{fig:6}.

The simplified transformer architecture, the direct MLP-style regression, and the GKP and KAR modules jointly constitute our Simple and strong Category-Agnostic Pose Estimator (SCAPE). The results on the MP-$100$ dataset~\cite{xu2022pose} indicate that, SCAPE invites $2.2$ and $1.3$ PCK improvements under the $1$-shot and the $5$-shot setting, respectively, with ResNet-50 backbone. With ViT backbones and strong pretraining, 
we observe significantly improved CAPE metrics, 
where SCAPE further reaches $91.95$  and $93.98$ PCK metric under the $1$-shot and the $5$-shot setting, with more pronounced relative improvement compared with the state of the art. Excluding the backbone, SCAPE uses only $51\%$ parameters compared with CapeFormer~\cite{shi2023matching}. Ablation studies are also conducted to justify our propositions. In a nutshell, sufficient evidence has been provided to show that SCAPE is a better baseline for CAPE.

\section{Related Work}
\label{sec:Related Work}

\subsection{Category-Agnostic Pose Estimation}
Category-Specific Pose Estimation (CSPE) has been the focus of the field for decades~\cite{newell2016stacked,xiao2018simple,sun2019deep,xu2022vitpose,song2019apollocar3d,reddy2018carfusion,li1906atrw,labuguen2021macaquepose}. Yet, estimating pose on new categories often means 
model redesign and retraining. Recently, the task of CAPE is introduced~\cite{xu2022pose}, and two baseline approaches POMNet~\cite{xu2022pose} and Capeformer~\cite{shi2023matching} are also proposed, enabling pose estimation on new categories with few exemplar images. 
The difference against CSPE is that the two approaches 
both focus on similarity metrics: keypoints are obtained by comparing the similarity between the query and support features. POMNet concatenates keypoint tokens with the query image, measuring similarity using a window to generate a similarity map, indirectly predicting coordinates from the map. 
Capeformer extends this approach into a two-stage framework, introducing a second stage to iteratively refine the unreliable matches from the first stage. This two-stage framework significantly enhances accuracy, but it is computationally expensive and adds $6.4\times$ more parameters than POMNet (excluding the backbone).
Our first goal is to reduce the computational burden in both model capacity and computation, without hurting accuracy.

\subsection{Similarity Matching}
Similarity matching is widely used in vision tasks like object detection~\cite{chen2016monocular,chen20153d} and tracking~\cite{bertinetto2016fully}, formulated as a problem of template matching. 
CAPE can also be viewed as executing similarity matching by aligning sparse support keypoints to the most similar keypoints in the target image. Semantic correspondence have typically explored point-level similarity matching using matching heads. However, challenges arise with models like SuperGlue~\cite{sarlin2020superglue} and SCOT~\cite{liu2020semantic}, indicating that relying solely on matching may lead to multiple responses in similarity maps, thus requiring additional post-processing modules like optimal transport for refinement. In the sparse setting of CAPE, CapeFormer~\cite{shi2023matching} uses a two-stage 
process to optimize matching results. With the rise of transformer, self-attention 
shows strong implicit matching capabilities. Recent studies in semantic correspondence has shifted towards transformer-based implicit matching, as seen in COTR~\cite{jiang2021cotr} and ACTR~\cite{sun2023correspondence}, which uses transformers for feature matching and fusion, without post-processing. However, sparse point semantic matching in the context of CAPE remains underexplored. This work focuses on leveraging implicit matching abilities of transformer for accurate CAPE.

\subsection{Keypoint Correlation in Pose Estimation}

In CSPE, such as in human pose estimation ,the correlation among keypoints has been proven to be a useful cue. 
Previous CNN-based models like ContextPose~\cite{ma2021context} have developed modules to model the keypoint correlation. Transformer-based approaches~\cite{shi2022end,geng2021bottom,li2021tokenpose} also reveal the implicit expression of keypoint correlation within them. These approaches treat keypoints as learnable tokens, akin to DETR~\cite{carion2020end}, allowing them to learn a universal representation for specific keypoints. The prior correlation among keypoints is also captured. 
In this way, these transformer models can easily capture correlations through attention maps between keypoints. However, keypoint correlation has yet to be exploited in CAPE. Inspired by the good practices in CSPE, we attempt to find a way to model and exploit keypoint correlation in CAPE.

\section{Simple and Strong Category-Agnostic Pose Estimator}
Here we present our SCAPE. We first 
review the framework and then reveal our explorations and understanding of the model and task. These explorations further motivate us to devise two enhancements specifically designed for CAPE.  
\begin{figure}[!t]
\centering
{\includegraphics[width=0.95\linewidth]{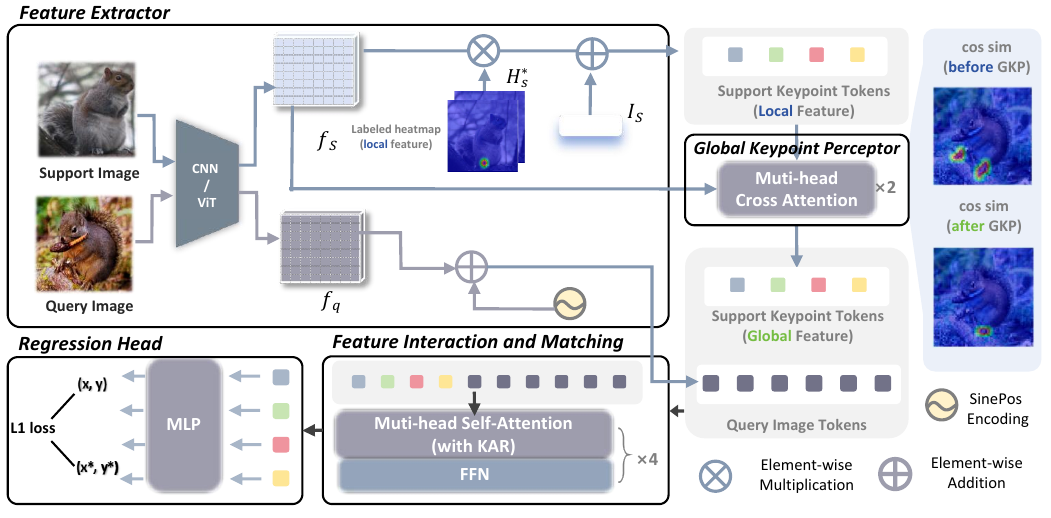}}
    \caption{\textbf{Technical pipeline of SCAPE.} SCAPE consists of four modules: feature extractor, global keypoint feature perceptor, feature interactor, and regression head. The feature extractor is identical to the poir method. 
    After processing support and query features through the backbone, support keypoint tokens are created by a weighted sum of the labeled keypoint heatmap and support features. Support keypoints (\textcolor{blue}{local}) are then combined with a support keypoint identifier (form Capeformer), while query image tokens are formed from the query features with positional embedding. Support keypoints feed in Global Keypoint Perceptor to cross-attend the support image to obtain \textcolor{green}{global} support keypoints. Next, an interaction module composed of multi-head refined self-attention (MHRSA) conduct the interaction between the keypoint and query features. And the Keypoint Attention Refiner (KAR) is inserted into each self-attention stage to refine the attention maps among keypoints. Finally, a simple MLP head are used for support keypoint to regress keypoint coordinates. 
    }
\label{fig:2}
\end{figure}
\subsection{Overall Framework}
Previous methods~\cite{xu2022pose,shi2023matching} interact features and measure the similarity between the query feature $F_q\in \mathbb{R}^{n \times c}$ and support keypoints $F_s\in \mathbb{R}^{k \times c}$ by concatenating them and feed into the convolutional layer to obtain a similarity map $S\in \mathbb{R}^{k \times n}$, then indirectly identifying the target keypoints $P\in \mathbb{R}^{k \times 2}$.
POMNet~\cite{xu2022pose} uses a decoder $\theta_M$ 
to infer similarity maps from combined support keypoints and query features, \ie, $S=\theta_M$ $(F_s,F_q)$. 
The keypoint coordinates are obtained by searching the peaks such that $P=\arg\max(S)$. This matching head accounts for $70$\% of the overall training memory consumption. 
CapeFormer~\cite{shi2023matching} simplifies 
this by taking the inner product of $F_q$ and $F_s$ after mapping. Both methods compare $F_s$ and $F_q$ to obtain similarity maps with supervision, categorized as similarity-explicit matching heads. CapeFormer uses a two-stage paradigm, the inferred coordinates $P_0$ are considered the initial coordinates. Then $P_0$ is refined at the second stage with offsets by $P^{l+1}=P^l+{\tt MLP}(F_s^{l+1})$. 
Our SCAPE adheres to the one-stage paradigm, but discards explicit similarity matching, and uses pure self-attention for implicit similarity matching and MLP direct coordinate regression. To enhance implicit matching in the transformer, we use GKP to enrich the semantics of $F_s$, reducing subsequent matching stress. We employ KAR to establish strong relations between keypoints, facilitating mutual assistance in point prediction. The technical pipeline of SCAPE model is shown in Fig.~\ref{fig:2}. We visualize the cosine similarity between the left foot keypoint and the support image (values less than 0.6 are masked) before and after the Global Keypoint Feature Perceptor (GKP) operation. After GKP, the similarity map of support keypoints (left foot), fused with global information, exhibits reduced responses compared to symmetric points. This reduction can be considered as alleviating subsequent matching pressure.

\subsection{Explicit Matching Is not Necessary}
Previous methods 
generate similarity maps and supervise them in map level, 
we refer to this operation as explicit matching. 
We first study the feature interactor of CapeFormer~\cite{shi2023matching} (all other implementation details stay fixed). As shown in Fig.~\ref{fig:cross_attention}, its similarity map has multiple responses, and the peaks do not converge. Yet, we observe that the final attention map of the feature interactor between support keypoints and query features is an effective similarity map, with clearer peaks. This suggests that the implicit similarity within self-attention seems a better substitution for explicit similarity. We surmise explicit matching has the following drawbacks: (i) 
Explicit matching methods, such as generating similarity maps and supervising on map level can lead to overfitting and increased matching difficulty, and (ii) it introduces extra blocks or modules that increase the complexity and bring expensive computation. We thus discard explicit similarity matching in CapeFormer, relying solely on the inherently implicit matching mechanism of attention, and also try direct coordinate regression from $F_s$,
\begin{equation}\small
    \label{eq:bbox}
    P={\tt MLP} ( F_s )\,.
\end{equation}
CapeFormer incorporates multiple loss functions, yet we solely supervise the coordinates $P$ output by the MLP, as indicated in Table~\ref{table:2}. Subsequently, we employ $\ell_1$ loss to compute loss. 
Such simple modifications lead us to a promising PCK of \textcolor{red}{$88.6$} on the split1 of the MP-100 dataset~\cite{xu2022pose}.

\begin{table}[!t]\small
\centering
\caption{Comparison with prior CAPE methods. `S-map' is the similarity map, and `coord' means direct coordinate regression, `offset' is supervise layer by layer.}
{
 \begin{tabular}{@{}l|c|cc|ccc|c@{}}
  \toprule
  & \multirow{2}{*}{stage}  & \multicolumn{2}{c|}{matching form} &  \multicolumn{3}{c|}{surpervison}&\multirow{2}{*}{end-to-end}\\
 && explicit & implicit&  s-map &coord&offset   &
\\
\midrule
POMNet &one& $\checkmark$ & &$\checkmark$ & & & \\
CapeFormer &two& $\checkmark$& & $\checkmark$& $\checkmark$& $\checkmark$&  \\
SCAPE & one& &$\checkmark$ & &$\checkmark$& &$\checkmark$ \\

\bottomrule
 \end{tabular}
}\label{table:2}
\end{table}


In addition, 
previous methods use DETR-Decoder for feature interaction, which we consider as a rough utilization of the DETR. Unlike the previous DETR-Decoder, which only updated $F_s$, our approach concatenates $F_s$ and $F_q$ and feeds them into self-attention, enabling simultaneous updates to both $F_s$ and $F_q$. This interpretation inspires us to replace the previous DETR-Decoder with self-attention layers for feature interaction. Therefore, our feature interaction, as shown in Fig.~\ref{fig:2}, concatenates $F_s$ and $F_q$, sending them into the fully self-attention Encoder for simultaneous updates. By doing so, we not only reduce the parameters but also improve accuracy to $89.1$ (\textcolor{red}{+$0.5$}).

In previous work, $F_s$ and $F_q$ possess distinct additional encodings. $F_s$ encompasses keypoint identifiers to differentiate between keypoints, while $F_q$ contains positional encoding. 
Using the same query and key weights for $F_s$ and $F_q$ is inappropriate, 
we therefore choose to have unshared projection weights for them. Therefore, in self-attention: ${K_s}$= $W_{K1}F_s$, ${K_q}$= $W_{K2}F_q$, ${K}$=$\tt{cat}$$(K_s,K_q)$, ${Q_s}$= $W_{Q1}F_s$, ${Q_q}$= $W_{Q2}F_q$, ${Q}$= $\tt{cat}$ $(Q_s,Q_q)$. 
The PCK further reaches $89.8$ (\textcolor{red}{+$0.7$}). 

\subsection{Global Keypoint Feature Perceptor}
\label{subsec:kcr}
Previous approaches initialize the support keypoints $F_s$ with only local features, \textit{i.e.}, extracting feature tokens from the support image centered on support ground-truth keypoints with Gaussian kernels. The extracted $F_s$ lacks global context, which hinders subsequent matching. As shown in Fig.~\ref{fig:6}, the attention process can not distinguish similar keypoints, leading to predictions shifting onto neighboring keypoints.
To address this, we further introduce Global Keypoint Feature Perceptor (GKP) that aims to infuse the global and context information from the support image to enrich the support keypoint representation. A cross-attention block where support keypoint features $F_s$ (query) cross-attend the support image enables this fusion process. Note that the introduction of GKP makes the structure more simple. In practice, the first two self-attention blocks are replaced by GKP. This replacement reduces both parameter count and matching complexity, leading to improved performance of 90.3 (\textcolor{red}{$+0.5$}) and accelerated model convergence.

\subsection{Keypoint Attenion Refiner}

\begin{figure}[!t]
\centering{\includegraphics[width=0.8\linewidth]{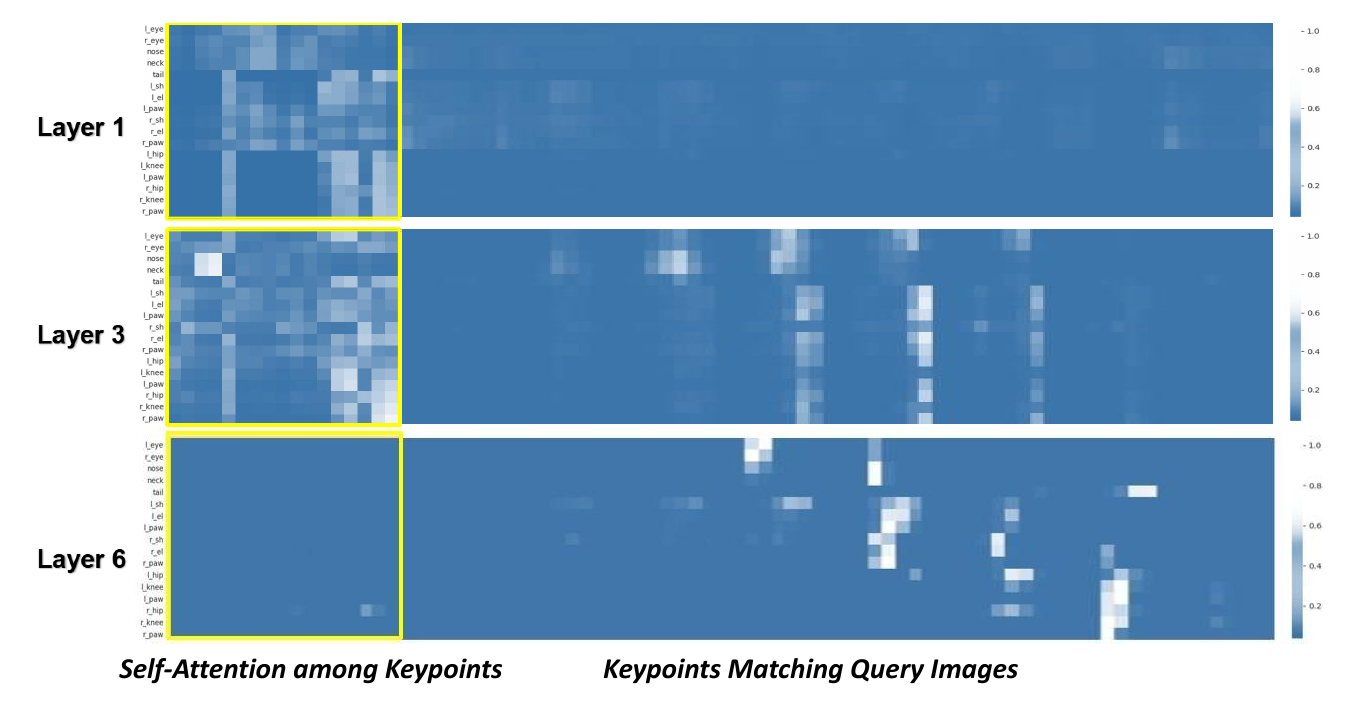}}
    \caption{\textbf{Visualization of the attention maps of support keypoints.} 
    Initially, due to significant disparities between support and query image, accurate matching is challenging, the attention process focuses on self-attention among keypoints (with \textcolor{yellow}{yellow} highlight) to build contextual information. Later attention leans towards implicit matching between keypoints to query images.}
\label{fig:3}
\end{figure}


\begin{figure}
\centering{\includegraphics[width=0.95\linewidth]{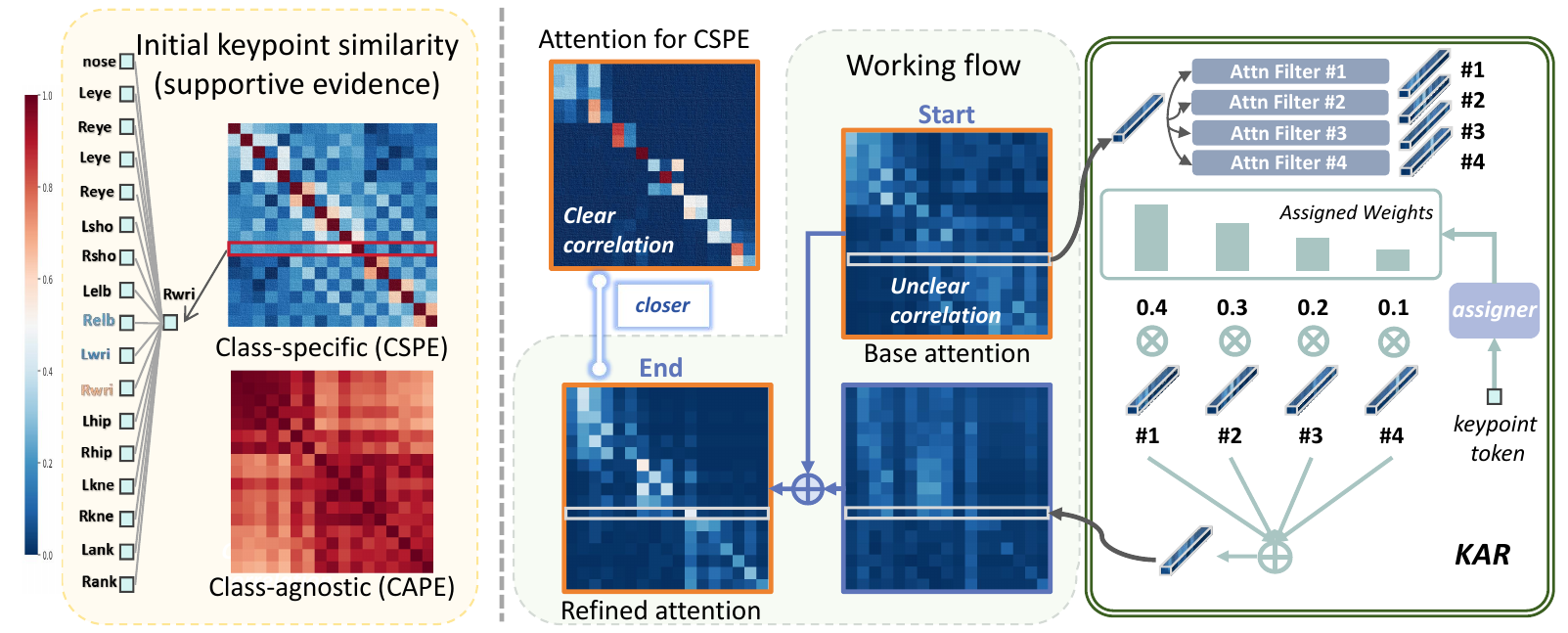}}
    \caption{\textbf{Attention map modulated by KAR 
    exhibits a clearer correlation.} The inner product (use cosine similarity with the range of $[0,1]$ and no row normalization) among initial keypoint tokens represents the prior keypoint correlation. In CSPE, such prior correlation is used as adjacency and symmetry constraints (strongly correlated with the right wrist are the left wrist and right arm), and transformers can easily model keypoint correlation (attention map) from these priors. And this correlation will aid keypoints in better locating each other. In CAPE, the unclear definition of prior correlation leads to noisy attention maps. KAR uses attention filters to filter out the noise of the base attention map and weights the output through a support keypoint weight assigner. The combined output with base attention yields a refined attention map, closer to the attention in CSPE.  } 
\label{fig:4}
\end{figure}
As shown in Fig.~\ref{fig:3}, in the early attention blocks, due to the different feature distributions, the attention stage concentrates on establishing the correlation among keypoints. As the number of layers increases, attention gradually shifts towards matching the query image. We aim to investigate the effectiveness of early-stage modeling of correlations among keypoints, so we mask the self-attention among keypoints (highlighted in \textcolor{yellow}{yellow}) and observe the PCK drop by \textcolor{green}{-1}. 
We argue that modeling keypoint correlation 
matters in CAPE.

However, in Fig.~\ref{fig:3}, We find attention maps among support keypoints show meaningless patterns, appearing to be random. However, transformer-based human pose estimation~\cite{li2021tokenpose} suggests that the attention among keypoints tends to be highly regular, showing certain correlation among keypoints. 
The specific keypoint tokens serve as learnable parameters~\cite{geng2021bottom,li2021tokenpose}. As shown in Fig.~\ref{fig:4}, 
data-driven learning tokens establish meaningful prior keypoint correlation (initial keypoint similarity) with other keypoints, enabling subsequent transformers to easily model reasonable keypoint correlation (attention map). In CAPE, due to unknown categories, the keypoint correlation in CAPE is difficult to model accurately by the transformer, thus introducing noise into the attention map of keypoints. We want to strengthen the correlation between keypoints by refining this attention map in self-attention (Encoder) blocks.

Firstly, we introduce the Attention Filter ($\tt AF$) to filter out noise. We set an attention filter, composed of a ReLU layer followed by an MLP, to filter out the undesired parts of the current base attention map ($\mathcal A$) and thus make the attention weights more distinct. The ReLU layer mainly charges the filtering, and the MLP re-models these correlations. The positive output signifies important node information, while unimportant details are filtered out. In this way, the PCK further reaches $90.9$ (\textcolor{red}{$+0.6$}) on split1. The filter can be written as:
\begin{equation}\small
    \label{eq:bbox}
     {\tt AF}(\mathcal{A})={\tt MLP}({\tt ReLU}(\mathcal{A}))\,.
\end{equation}
Since keypoints of different categories may benefit from distinct structural data propagation, we further introduce the Keypoint Attention Refiner (KAR). KAR incorporates multiple keypoint filters to enhance the modeling of structural details. It processes support keypoints $F_s$ through a keypoint weight assigner ($\tt Assign$) to determine the weights of different keypoint filters, and one can observe that similar tokens (dog leg and cat leg) possess similar weights (Per supplementary materials). 
Formally, KAR is defined by:
\begin{equation}\small
    \label{eq:bbox}
     {\tt KAR} (\mathcal{A} )=\sum_{i=1}^{n(n=4)} {{\tt Assign}_i(F_s)}{\tt AF}_i(\mathcal{A})\,,
\end{equation}
where $\tt Assign_i$ performs a linear projection on $F_s$ by a weight matrix $W_{\tt Assign}$, with dropout and layer normalization within $\tt Assign_i$ to prevent overfitting. Then, a softmax function is applied 
for another normalization, defined by
\begin{equation}\small
    \label{eq:bbox}
     {\tt Assign}(F_s)={\tt softmax}({\tt layernorm}(({\tt dropout}(F_sW_{\tt assign})))\,,
\end{equation}
The output of KAR 
is summed with the base attention map $\mathcal{A}$ to yield the refined attention $\mathcal{A}_{\tt refined}$
\begin{equation}\small
    \label{eq:bbox}
     \tt  \mathcal{A}_{refined}=softmax(\mathcal{A}+ KAP(\mathcal{A}))\,.
\end{equation}
The refined attention, as shown in Fig.~\ref{fig:4}, shows clearer keypoint correlation. By collaborating with multiple $\tt AF$, we further advance PCK to $91.6$ (\textcolor{red}{$+0.7$}). In addition, previous methods removed~\cite{shi2023matching} support keypoint identifier led to a significant performance decrease. However, adding the KAR module and then removing $I_s$ resulted in a slight decrease.

\section{Experiments}

\subsection{Implementation Details}
\label{subsec:implementation_detail}
\textbf{Dataset.} We use the MP-100 dataset provided by Xu \etal~\cite{xu2022pose}. Being the first large-scale dataset for CAPE, the MP-100 dataset consists of $100$ object categories and contains over $20,000$ instances.

\noindent \textbf{Data Pre-processing.} Following POMNet~\cite{xu2022pose}, we divide the data into $5$ splits, each of which includes $100$ categories, among which $70$ for training, $10$ for validation, and $20$ for testing, ensuring no overlap between categories. We apply the same data pre-processing as POMNet. 

\noindent \textbf{Metric.} PCK (Probability of Correct Keypoint) is a widely used metric for evaluating pose estimation algorithms. Like POMNet, we use PCK as the quantitative metric and set the threshold to $0.2$ for all categories, and the mean of PCK across $5$ splits is also reported. To address the limitation of PCK, we further report AUC~\cite{khan2017synergy} and NME~\cite{dong2018supervision} to evaluate the localization performance. Please refer to the supplementary material for additional details.

\noindent \textbf{Network Architecture.} To validate the adaptation of our approach across different backbones, we choose: 1) ResNet-$50$~\cite{he2016deep}, 2) ViT-B~\cite{dosovitskiy2020image} and Swin-S~\cite{liu2021swin} pretrained on ImageNet-1K~\cite{deng2009imagenet}, and 3) ViT-S and ViT-B trained with DINOv2~\cite{oquab2023dinov2}. Like CapeFormer~\cite{shi2023matching}, both the support and the query image share the same feature extractor. Then, We first uses $2$ Global Keypoint Feature Perceptor (cross-attention), followed by $4$ Feature Interactor (self-attention), termed SCAPE.
 To showcase the effectiveness of our approach, we streamline the model by introducing \textbf{Lite-SCAPE}, a baseline with only 3 attention blocks (1 GKP and 2 feature interactors). The parameter for Lite-SCAPE is provided below, and additional performance details can be found in the supplementary material.

\noindent \textbf{Training Details.} Following POMNet, we apply the Adam optimizer~\cite{kingma2014adam} with a batch size of $16$. The model is trained for \textcolor{red}{$180$} epochs with an initial learning rate of $2e^{-4}$, while in pervious method~\cite{xu2022pose}~\cite{shi2023matching} the totol epoch is \textcolor{red}{$210$}. The learning rate is multiplied by $0.1$ at the $140$-th and the $170$-th epoch (cosine annealing can achieve the same effect), respectively. We solely employ $\ell_1$ loss to supervise the regression of 2D coordinates.

\begin{figure*}[!t]
\centering{\includegraphics[width=1.0\linewidth]{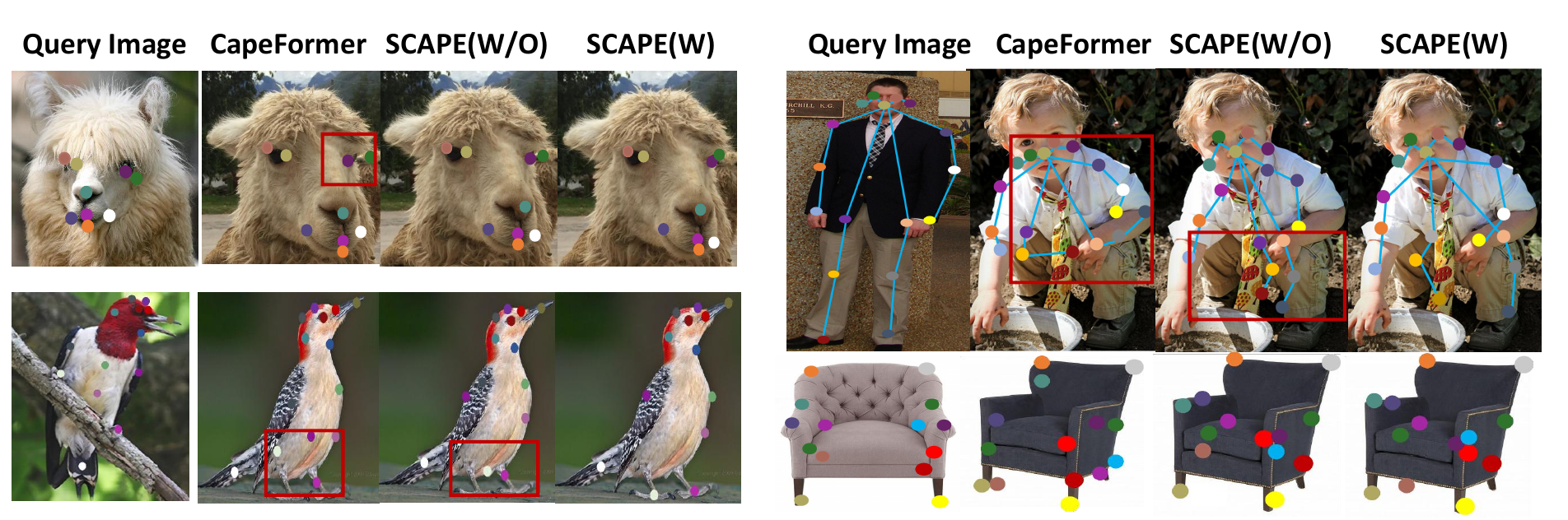}}
    \caption{ 
Visual results of CapeFormer, SCAPE (W/O KAR and GKP) and SCAPE(W KAR and GKP) on the MP-100 dataset. \textcolor{red}{Red boxes} indicate incorrect estimates.}
\label{fig:last}
\end{figure*}

\begin{table}[!t]
 \caption{Comparison with state-of-the-art approaches on the MP-100 dateset under the 1 and 5-shot setting. Best performance is in \textbf{boldface}.}
\scriptsize
\centering
\renewcommand{\arraystretch}{1.1}
\addtolength{\tabcolsep}{2pt}
 {
\begin{tabular}{@{}cc|ccccc|c@{}}
\toprule
1-shot &  Backbone  &split1&split2  &split3 &split4  &split5 & Mean \\ \hline
ProtoNet~\cite{lyu2019advances}  &R50&  46.05 &40.84 &49.13  &43.34 &44.54 &44.78    \\
MAML~\cite{finn2017model}      &R50& 68.14 &54.72 &64.19 &63.24 &57.20  &61.50    \\
Fine-tune~\cite{nakamura2019revisiting} &R50& 70.60 &57.04 &66.06  &65.00 &59.20  &63.58    \\
POMNet~\cite{xu2022pose}     &R50&  84.23&78.25 &78.17 &78.68 &79.17  &79.70
    \\
CapeFormer~\cite{shi2023matching}&R50& 89.45& 84.88& 83.59  & 83.53  &85.09 &85.31 \\
SCAPE (Ours)  &R50&   \textbf{91.67}& \textbf{86.87}&\textbf{87.29} & \textbf{85.01} &\textbf{86.92} &\textbf{87.55} \\ 
\hline
\\
\hline
5-shot &  Backbone  &split1&split2  &split3 &split4  &split5 & Mean \\ \hline
ProtoNet~\cite{lyu2019advances}  &R50& 60.31 &53.51 &49.13  &43.34 &44.54  &44.78     \\
MAML~\cite{finn2017model}      &R50& 68.14 &54.72 &64.19  &63.24 &57.20  &61.50    \\
Fine-tune~\cite{nakamura2019revisiting} &R50& 70.60 &57.84&66.76  &66.53 &60.24  &64.61    \\
POMNet~\cite{xu2022pose}     &R50&  84.72&79.61 &78.00 &80.38 &80.85  &80.71
    \\
CapeFormer~\cite{shi2023matching}&R50&91.94& 88.92& 89.40  & 88.01  &88.25 &89.30 \\
SCAPE (Ours)  &R50&   \textbf{93.42}& \textbf{89.91}& \textbf{90.61} & \textbf{89.44 }&\textbf{89.95}&\textbf{90.66}  \\ 
\bottomrule

\end{tabular}
}

\label{tab:3}
\end{table}

\subsection{Results on the MP-100 Dataset}
We first compare the performance 
to show the effectiveness and efficiency of our approach. Following previous methods, we test our model on MP-100 dataset.
\begin{table}[!t]
 \caption{Performance across different transformer backbones. We report the mean across 5 splits on the MP-100 dataset. The improvement relative to R50 as the backbone is highlighted in \textcolor{red}{red}.}
\scriptsize
\centering
\renewcommand{\arraystretch}{1.2}
\addtolength{\tabcolsep}{3pt}
{
\begin{tabular}{@{}cccccc@{}}
\toprule

Method & Backbone & 1-shot Mean & 5-shot Mean\\ 
\midrule
SCAPE(Ours)& ViT-B & 88.19  & 91.53\\
SCAPE(Ours)&  Swin-S &87.93 & 92.01\\
SCAPE(Ours)&  ViT-S (DINOv2) &90.74& 92.65\\
\midrule
CapeFormer & ViT-B (DINOv2) & 89.11\textcolor{red}{(+3.8)} &  91.91\textcolor{red}{(+2.6)} \\
SCAPE(Ours)&  ViT-B (DINOv2)  & \textbf{91.95}\textcolor{red}{(+4.4)} & \textbf{93.98}\textcolor{red}{(+3.4)} \\
\bottomrule
\end{tabular}
}
\label{tab:dif}
\end{table}

\noindent \textbf{Effectiveness.}
As shown in Table~\ref{tab:3}, SCAPE outperforms other approaches on all splits. Compared with CapeFormer, using the same R50 as the backbone, SCAPE achieves an improvement of $+2.2$ and $+1.3$ under the $1$-shot and the $5$-shot setting respectively. Qualitative results are shown in Fig.~\ref{fig:last}. 
Additionally, our simple framework is highly compatible with transformer-based backbones. In Table~\ref{tab:dif}, To underscore our high compatibility, we compare SCAPE with CapeFormer using a DINOv2-pretrained VIT-B approach, achieving improvements of +$4.4$ and +$3.4$, notably surpassing the performance gap on R50. 
Note that, SCAPE with ViT-S even outperforms CapeFormer ViT-B. This reveals the benefit of a simple end-to-end architecture for CAPE.

\begin{table}[!t]
\caption{Efficiency comparison are tested on $256\times256\times3$ input on RTX 3090 
The experiments are conducted under the 1-shot setting of the MP-100 dataset. R50 is employed as backbone. 
$\Delta$Params indicates the model parameters (excluding backbone).}

\scriptsize
\centering
\renewcommand{\arraystretch}{1.1}
\addtolength{\tabcolsep}{2pt}
{
\begin{tabular}{@{}cccccccc@{}}
\toprule
Method    &Attn Blocks&  GFLOPs  & $\Delta$Params (M) & Mem (G) & FPS & PCK Mean \\ 
\midrule
POMNet &6 &38.01 &+1.19   &13.8*2     &6.80 &79.70\\
CapeFormer&9 &23.68 &+7.63   &7.8     &26.09&85.31\\
Lite-SCAPE(ours) &3 &22.20  &+1.95  &6.0  &36.89  &86.13\\
SCAPE(ours) &6 &22.81  &+3.88  &6.3  &29.43  &87.33\\
\bottomrule
\end{tabular}
}

\label{tab:5}
\end{table}

\begin{table}[!t]
\caption{Cross super-category pose estimation 
under the $1$-shot setting.}
\scriptsize
\centering
\renewcommand{\arraystretch}{1.1}
\addtolength{\tabcolsep}{2pt}
{
\begin{tabular}{@{}cccccc@{}}
\toprule
Method     & Human Body &Human face    &Vehicle &Furniture\\ 
\midrule

ProtoNet& 37.61 &57.80 &28.35 & 42.64\\
MAML  &51.93 &25.72 &17.68 &20.09 
   \\
Fine-tune &52.11  &25.53 &17.46 &20.76   \\

POMNet &73.82  &79.63 &34.92 &47.27   \\
CapeFormer &83.44  &80.96 &45.40 &52.49    \\
SCAPE(ours)  &\textbf{84.24}  &\textbf{85.98} &\textbf{45.61}&\textbf{54.13}\\
\bottomrule
\end{tabular}
}
\label{tab:cross}

\end{table}

\noindent \textbf{Efficiency.}
From Table~\ref{tab:5}, Here we provide more evidence to show that SCAPE achieves a good trade-off between efficiency and performance, further enhancing model performance without much workload. Note that, our light version Lite-SCAPE (with only half blocks of SCAPE and 1/3 of CapeFormer) still outperforms CapeFormer by $+0.8$ PCK, with reduced additional parameters by $75$\%, boosting the inference speed by $39$\%. Our simplicity is not only evident in our design but also reflected in practical efficiency.

\begin{table}[!t]\small
\centering
\renewcommand\arraystretch{1}
\addtolength{\tabcolsep}{-1.5pt}
\end{table}

\subsection{Cross Super-Category Generalization}

Previous experiments suggest limited generalization ability, meaning the capacity to predict specific categories heavily relies on knowledge acquired from those categories. To validate the generalization of our model, we conduct a cross super-category experiment, similar to previous methods. As shown in Table~\ref{tab:cross}. ours exhibits good generalization ability, 
yielding best performance. 
\subsection{Ablation Study}
Here we verify our design choices: i) the comparison between the regression head and the matching head, along with supervision signal choices; ii) the design choice of KAR; iii) the effectiveness of individual components in SCAPE; iv) the number of attention blocks and the proportion of attention layers used by GKP and feature interaction. All experiments are conducted under the 1-shot setting on split1. And the design in \textcolor{blue!31}{light blue} is our choice.

\noindent \textbf{Matching and Regression.}
Our approach relies on implicit self-attention matching to directly regress coordinates. This is different from the explicit matching used in previous methods that indirectly obtain coordinates from similarity maps. We justify two main differences: 1) the form of matching (implicit vs.\ explicit), and 2) the form of supervision signal or get keypoints (coordinates vs regressed similarity maps).
Results are shown in Table~\ref{tab:s6}. For 1), We use L1 (Line 1) as a baseline which applies the additional matching head and map-level supervision on SCAPE (w/o GKP and KAR). L2 directly regresses similarity maps from the final keypoint tokens and outperforms L1, which shows that implicit matching proves superiority over explicit matching. For 2), we compare L2 and L3, and the results demonstrate that direct regression of coordinates outperforms regressing similarity maps.
\\ \hspace*{\fill} \\
\begin{minipage}{\textwidth}
\begin{minipage}[!t]{0.48\textwidth}
\makeatletter\def\@captype{table}
\scriptsize
\centering
\caption{Validation of matching head and direct regression head.}
{
\begin{tabular}{ccc}
\toprule
 Paradigm  &Result Form& PCK \\ 
 \midrule
Matching&  similarity map&$86.3$  \\
 Regression&  similarity map&$88.2$ \\
\rowcolor{blue!5}  Regression&  coordinate&$89.1$ \\
\bottomrule
\end{tabular}
}
\label{tab:s6}
\centering
\label{sample-table}
\end{minipage}
\begin{minipage}[!t]{0.48\textwidth}
\centering

\makeatletter\def\@captype{table}
\caption{The design of multiple Attention Filters in KAR.}
\scriptsize
\renewcommand{\arraystretch}{1.2}

{
 {
 \begin{tabular}{@{}lccc@{}}
 \toprule
  $\tt AF$&hidden-dim&PCK 
\\ \hline
 $1$&$50$ &$90.5$ \\
$1$ &$200$&$90.3$  \\
\rowcolor{blue!5}$4$ &$50$& $91.2$ \\
\bottomrule
 \end{tabular}
}
\label{tab:s7}

}

\label{sample-table}

\end{minipage}
\end{minipage}
\\ \hspace*{\fill} \\

\noindent \textbf{Design of Keypoint Attention Refiner (KAR).}
We introduce to employ multiple(4) Attention Filters to refine the base attention simultaneously. 
This design can further tackle the varying categories for CAPE (PCK \textcolor{red}{$+0.7$}). To figure out whether the improvement comes from the increased parameters or the multiple aspects modulation
of node relationships. 
We augment a single Attention Filter by increasing its hidden layer dimensions to n times the original, to align with parameters of n Attention Filters for a fair comparison. Specifically, the linear transformation changes from (100-50-100) to (100-200-100), with corresponding results presented in Table~\ref{tab:s7}. Surprisingly, increasing the hidden layer dimensions results in a performance drop of \textcolor{green}{-0.1}. 
On the contrary, increasing the number of Attention Filter can boost the performance, which suggests the improvement comes from better refinement to attention map but not parameters.

\noindent \textbf{Performance Gains of SCAPE Components.}
In Table~\ref{table:8}, the comparison between S1 and S2 indicates that the feature interaction of the full encoder outperforms DETR (used by CapeFormer) by $0.5$. By comparing S2 with S3, unsharing query and key weights in self-attention between keypoint tokens and query image leads to a performance gain of $0.7$. By contrasting S3 with S5, KAR shows an improvement of $+1.4$. By comparing S3 with S4, GKP demonstrates an improvement of $+1.0$, and combining these two modules proposed by us further enhances the performance.

\noindent \textbf{Performance of Different Attention Blocks.}
As shown in Table~\ref{tab:9}, employing $6$-layer attention blocks, involving $2$ layers of GKP to refine initial support keypoint tokens and $4$ layers of the self-attention feature interactor, exhibits the best performance. 
Additionally, increasing the number of attention blocks also improves performance. 

 \noindent \textbf{Further Exploration.} We address some uncertainties in the supplementary materials. (a) Our proposed GKP and KAR are tailored to the CAPE task, and we assess their generalization across other CAPE models. (b) Some existing large models exhibit class-agnostic matching and even multi-task capabilities, it would be interesting to evaluate whether CAPE-specific models are even needed.
\\ \hspace*{\fill} \\
 \begin{minipage}{\textwidth}
\begin{minipage}[t]{0.5\textwidth}

\makeatletter\def\@captype{table}
\caption{Performance gains of SCAPE components.}
\scriptsize
\centering
{
 \begin{tabular}{@{}l|cc|cc|c@{}}
  \toprule
    & \multicolumn{2}{c|}{Interactor} &  \multicolumn{2}{c|}{}\\
 & Form& Unshare Q/K&  KAR&GKP & PCK 
\\
\midrule
S$1$ &DETR& -- & & &88.6 \\
S$2$ & Encoder& & & & 89.1  \\
S$3$ & Encoder&$\checkmark$ & & &89.8 \\
\midrule
S$4$&Encoder & $\checkmark$& & $\checkmark$&90.8 \\
S$5$ & Encoder&$\checkmark$ &$\checkmark$ & &91.2  \\
S$6$ &Encoder&$\checkmark$ &$\checkmark$ &$\checkmark$ &91.9 \\
\bottomrule
 \end{tabular}
}

\centering
\label{table:8}

\end{minipage}
\begin{minipage}[t]{0.45\textwidth}

\renewcommand{\arraystretch}{1.2}
\centering
\makeatletter\def\@captype{table}
\scriptsize
\caption{Performance of proportion occupied by GKP and feature fusion layers and number of attention blocks.}
{
\begin{tabular}{cccc}
\toprule
 Proportions&GKP & Interactor  &PCK\\ 
\midrule
 All Interactor&$0$  & $6$ &$91.2$  \\

 1:1 & $3$  &$3$&$91.2$ \\
 \rowcolor{blue!5} 1:2 & $2$  & $4$ & $91.6$\\
 \midrule
 Lite-SCAPE & $1$ & $2$ & $90.4$ \\
 \rowcolor{blue!5} SCAPE & $2$ & $4$ & $91.6$ \\
\bottomrule

\end{tabular}
}

\label{tab:9}
\end{minipage}
\end{minipage}
\section{Conclusion and Limitaion}
\textbf{Conclusion} We propose a simple, strong, and straightforward CAPE baseline. Initially, we rely solely on implicit attention matching, discarding complex explicit matching heads, and directly regress coordinates with a simple MLP head. Subsequently, to enhance attention quality in CAPE, we introduce GKP to equip support keypoints with global semantics.
Then we present KAR to establish correlations among keypoints to enable the inference from related keypoints. 
SCAPE surpasses the state-of-the-art CAPE models in 
accuracy and efficiency.

\noindent\textbf{Limitation} We test the applicability of SCAPE on multiple instance scenarios across categories and even across domains, as per the supplementary materials. 
The current CAPE only finds one-to-one correspondence. 
We believe that future CAPE could explore multi-instance scenarios.

\noindent\textbf{Acknowledgement} This work is supported by Hubei Provincial Natural Science Foundation of China under Grant No. 2024AFB566.

\appendix

\section*{Appendix}
The appendix includes the following content: 
\begin{itemize}
    \item Additional results of SCAPE and on all splits of MP-$100$;
    \item More details about the proposed framework;
    \item Additional ablation experiments;
        \item Further Exploration for CAPE;
        \item The applicability of SCAPE on more scenarios.
        \item The code for SCAPE and and the testing for Painter~\cite{wang2023images} on MP100.
    \item More qualitative results;
    \item Visual comparison of attention maps between our state-of-the-art methods and ours.
    
\end{itemize}

\renewcommand\thesubsection{\Alph{subsection}}
\subsection{Additional results of SCAPE on
MP-100}

\vspace{5pt}
\noindent\textbf{Additional metrics reported.} As mentioned in Section $4.1$, PCK has some limitation on evaluating the performance. Here we report two additional metrics including Area Under the Curve (AUC)~\cite{khan2017synergy} and Normalized Mean Error (NME)~\cite{dong2018supervision} for CapeFormer and ours on the MP-100 dataset as shown in Table \ref{tab:s1}. 

\vspace{5pt}
\noindent\textbf{Complete results for Table 3 } We supplement the complete results of Table 3 in main content.
across five splits on both 1-shot and 5-shot setting on the MP-100 dataset as shown in Table \ref{tab:s2}.

\subsection{More details about the framework}
\vspace{5pt}
\noindent\textbf{Downsampling the feature output by transformer backbones. 
}
Due to the inconsistent resolution output by ResNet ($8\times8$) and transformer backbones ($16\times16$), we add an average pooling layer at the end of the used transformer backbones for workload reduction and fair comparison.

\vspace{5pt}
\noindent\textbf{The structural differences between SCAPE (without GKP and KAR) and SCAPE (with GKP and KAR).}
SCAPE (w/o GKP and KAR) consists of six layers of feature interaction. 
To enhance the quality of the attention map, we extends SCAPE by incorporating two designs, Global Keypoint Feature Perceptor (GKP) and Keypoint Attention Refiner (KAR). In practice, the first two self-attention blocks are replaced by Light GKP. The GKP is a cross-attention layer which refines the initial support keypoint tokens, with a total of $M=2$ layers in SCAPE. And the KAR is inserted into each feature interaction (self-attention) stage to refine the attention maps among keypoints. In a nut shall, to maintain the efficiency, we decrease the number of feature interaction layers $N=6$ in SCAPE (w/o GKP and KAR) to $N=4$ in SCAPESCAPE (w GKP and KAR). Moreover, the Lite-SCAPE model has $M=1$ GKP module and $N=2$ interactive modules.

\vspace{5pt}
\noindent\textbf{The results of Lite-SCAPE on the MP-100 dataset.
} Table~\ref{tab:s3} provides the full metrics as supplement for Table 4 in main content, where one can see that Lite-SCAPE performs well on all five splits of the MP-100 dataset. 

\begin{table}\small
 \renewcommand\arraystretch{1}
 \centering
 \addtolength{\tabcolsep}{6pt}
 \caption{comparison with the results of AUC($\uparrow$) and NME($\downarrow$) on $5$ splits under the $1$-shot setting of the MP-100 dataset. Best performance is in \textbf{boldface}.}
 {
\begin{tabular}{@{}lcccccccc@{}}
\hline
\toprule
method & metric & split1 &split2    &split3 &split4 &split5 & mean \\
\hline
\multirow{2}{*}{CapeFormer} & NME$\downarrow$ &0.088  &0.110 &0.111&0.116&0.108&0.106 \\
& AUC$\uparrow$ &88.64&86.39&86.18&85.81 &86.51&86.70 \\

\multirow{2}{*}{SCAPE}& NME$\downarrow$&\textbf{0.078}  &\textbf{0.101} &\textbf{0.097}&\textbf{0.105}&\textbf{0.101}&\textbf{0.096} \\
&  AUC$\uparrow$ &\textbf{89.59}&\textbf{87.12}&\textbf{87.34}&\textbf{86.28}&\textbf{86.96} &\textbf{87.45}
\\
\bottomrule
\hline
\end{tabular}
}
\label{tab:s1}
\vspace{-10pt}
\end{table}

 \begin{table}\footnotesize
 \renewcommand\arraystretch{1}
 \centering
 \caption{Performance across different Transformer-based backbones, we report all 5 splits on the MP-100 dataset, considering both 1-shot and 5-shot settings. Best performance is in \textbf{boldface}.}
 \addtolength{\tabcolsep}{4pt}
 {
\begin{tabular}{@{}lcccccccc@{}}
\hline
\toprule
method &  backbone & shot & split1&split2  &split3 &split4  &split5 & mean \\ \hline

\multirow{2}{*}{SCAPE} & \multirow{2}{*}{ViT-B} & $1$ & $91.74$&$87.57$ &$87.70$ &$86.49$ &$87.46$&$88.19$ \\
& & $5 $& $94.83$ & $90.65$ & $90.94$ &$90.98$ & $90.19$ & $91.52$ \\
\multirow{2}{*}{SCAPE}& \multirow{2}{*}{Swin-S} & 1 &$91.66$ & $87.01$&$86.98$&$85.97$&$87.91$ &$87.91$ \\
& & $5$ & $95.18$ & $91.25$ & $91.78$ & $90.74$ & $91.10$ & $92.01$ \\
\multirow{2}{*}{SCAPE}& ViT-S& 1 & $94.47$&$89.55$ &$89.81$&$89.04$ &$90.85$ &$90.74$ \\
& (DINOv2) & $5$ & $96.29$ & $92.11$ & $90.48$ & $92.27$ & $92.11$ & $92.65$\\
\midrule
\multirow{2}{*}{CapeFormer} & ViT-B& $1$ & $93.43$& $89.03$&$87.50$  & $86.32$  &$89.31$ &$89.11$  \\
& (DINOv2) &$5$ & $95.34$ & $92.10$ & $90.84$ & $90.60$ &$ 90.71$ & $91.92$ \\

\multirow{2}{*}{SCAPE} & ViT-B & 1 &  $\textbf{95.01}$ &$\textbf{90.65}$ &$\textbf{90.65}$&$90.50$&$\textbf{92.97}$ &$\textbf{91.95}$ \\
& (DINOv2) & 5 & $\textbf{97.10}$ & $\textbf{93.28}$ & $\textbf{92.02}$ &$\textbf{92.83}$ & $\textbf{94.67}$ & $\textbf{93.98}$ \\
\bottomrule
\hline
\end{tabular}
}

\label{tab:s2}

\end{table}

\begin{table}\small
\centering
\caption{Lite-SCAPE on the MP-100 dateset  on $5$ splits under the $1$-shot setting}\label{tab:s3}
\renewcommand\arraystretch{1}
\begin{tabular}{@{}lcccccc@{}}
\hline
\toprule

split1 & split2 &split3 &split4&split5& mean(PCK) 
\\ 
\midrule
 $90.01$&$85.17$&$85.45$&$84.91$&$85.14$&$86.13$\\

\bottomrule
\hline
\end{tabular}
\end{table}

\subsection{Additional Ablation Studies 
}

\vspace{5pt}
\noindent\textbf{The design of Global Keypoint Feature Perceptor.} In Section $4.4$, we focus on the design of GKP, as depicted in Table~\ref{tab:10}, GKP enriches the semantic content of the initial keypoint tokens by interacting again with the support image. In implementation of attention, the term "key" and "value" encompass not only support images but also include $F_q$ (query images). By comparing D$1$, D$2$ and D$3$, interacting the support image is essential with GKP and allowing the support keypoints to pre-examine the $F_q$ intended for fusion can further improve performance. Then, as the core of our GKP lies in interacting with the support image. We explore the possibility of incorporating this interaction into the following feature interactor. That is to say, in addition to concatenating $F_s$ and $F_q$ as before, we also concatenate the support image and update all three in feature interaction. According to Table~\ref{tab:11}, performance of the second line drops when incorporating the support image. The expected match for the feature interaction module is only the query image, and including support images disrupts matching. Therefore, the support image cannot be introduced during the feature interaction stage.

\begin{table}[t]\small
\centering
\caption{Design of Global Keypoint Feature Perceptor .}
\renewcommand\arraystretch{1}
{
 \begin{tabular}{@{}lccccc@{}}
   \toprule
&Support image&  Query image&  PCK 
\\
\midrule
D$1$ &$\checkmark$&  & & $90.0$ \\
D$2$ & & $\checkmark$& &  $89.3$  \\
\rowcolor{blue!5}D$3$ &$\checkmark$ &$\checkmark$ &$\checkmark$ &$90.3$ \\
\bottomrule
 \end{tabular}
}
\label{tab:10}

\end{table}

\begin{table}[!t]\small
\centering
\caption{The Global Keypoint Feature Perceptor must be a separate module and cannot be integrated into the feature fusion layer. }
\renewcommand\arraystretch{1}
{
 \begin{tabular}{@{}c|ccc@{}}
 \toprule
   KGP& Interactor &Support image  & PCK 
\\ 
\midrule
 0& 6 & &89.8 \\
0& 6 &$\checkmark$ &89.3  \\
\rowcolor{blue!5} 2&4 & &90.3 \\
\bottomrule
 \end{tabular}
}
\label{tab:11}
\end{table}

\begin{figure}
\centering{\includegraphics[width=0.35\linewidth]{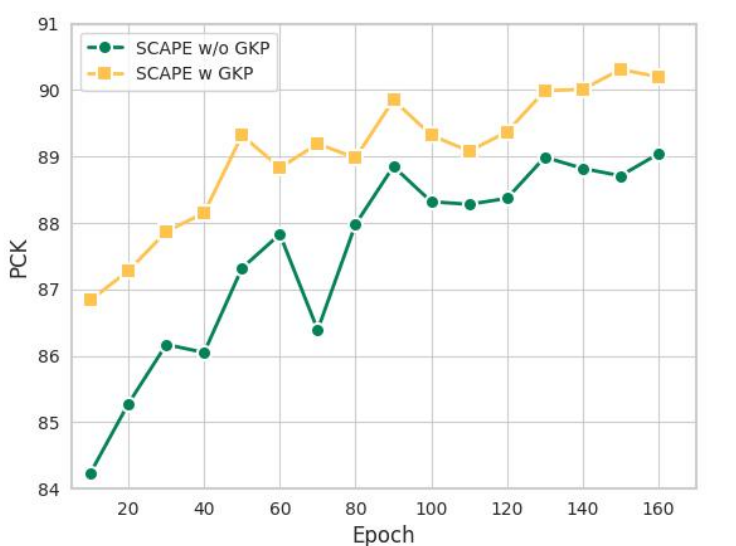}}
\caption{ \textbf{The result of SCAPE w/ and w/o the GKP.} We train the model for a total of 160 epochs under the $1$-shot setting on split-$1$ of MP-100, and evaluate the PCK metric every 10 epochs. The x-axis and y-axis represents the number of epochs and the PCK respectively.}
\label{fig:curve}
\end{figure}

\noindent\textbf{Global Keypoint Feature Perceptor can expedite convergence.} 
Learning curves in Fig.\ref{fig:curve} S1 illustrate that GKP accelerates convergence and enhances final performance. With the assistance of the GKP, SCAPE achieves optimal performance of 90.3 within 160 epochs, whereas previous methods were trained for 210 epochs.

\begin{figure}[!t]
\centering{\includegraphics[width=0.8\linewidth]{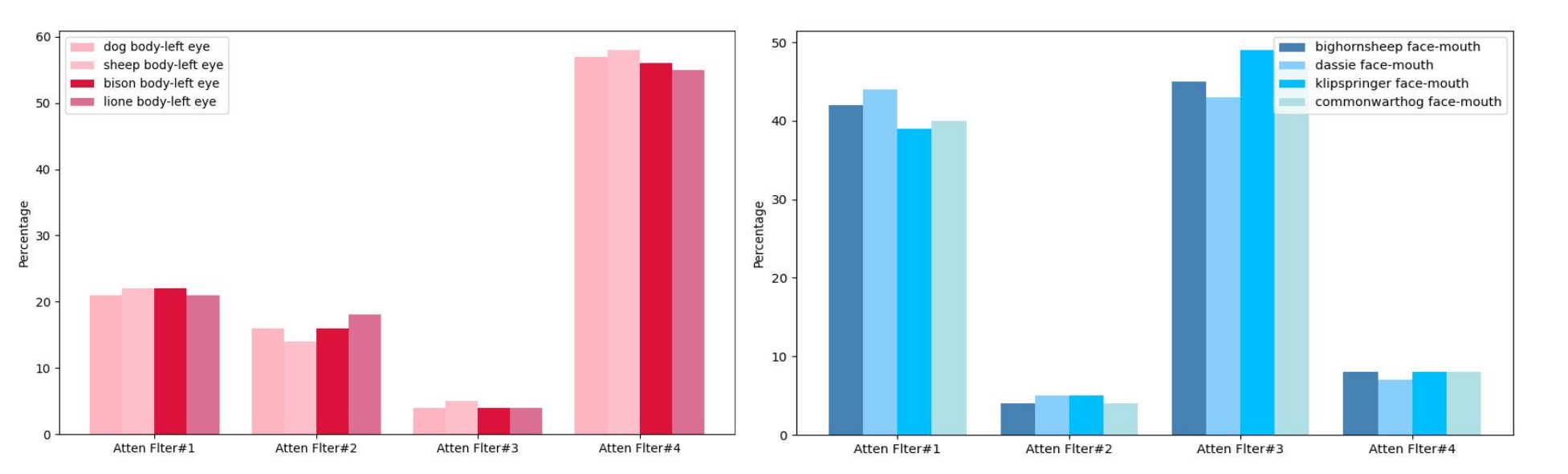}}
\caption{ \textbf{Similar  support keypoints exhibit analogous weight distributions (for the weight assigner).}  Even though they belong to different categories, their keypoints are the same, and the inter-node relationships is similar, resulting in closely aligned weight distributions.
}
\label{fig:KCR1}
\end{figure}

\begin{figure}[!t]
\centering{\includegraphics[width=0.8\linewidth]{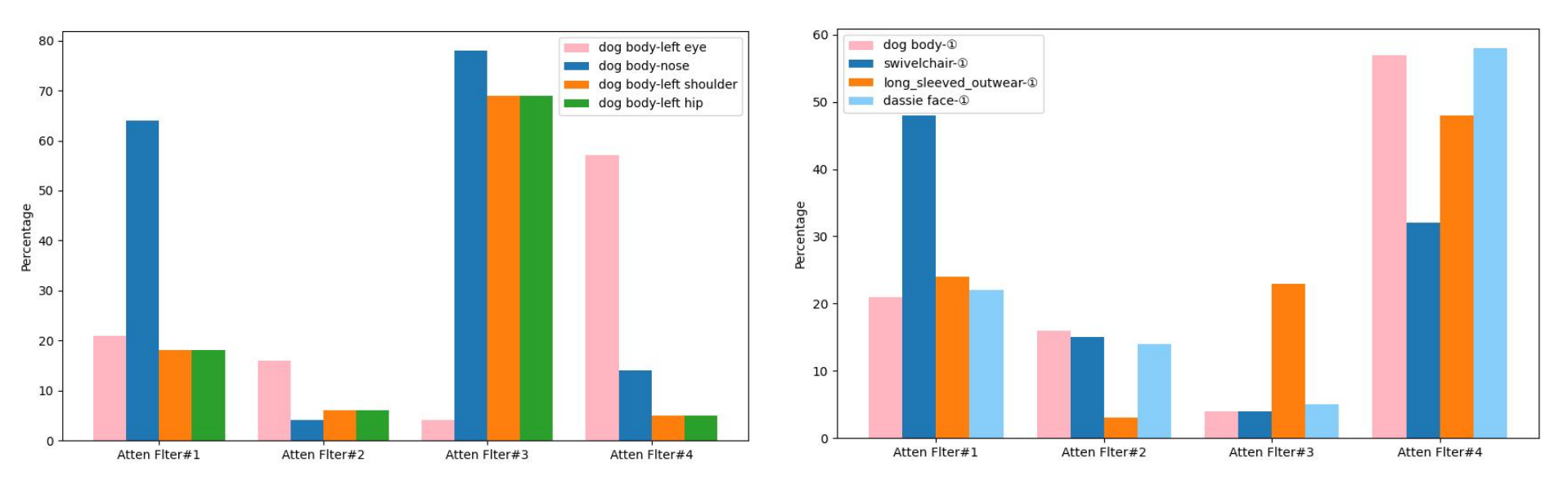}}
\caption{ \textbf{The weight distributions of dissimilar support keypoints differ.} On the left side, nodes within the same category exhibit varying distributions due to different ways of modeling inter-node relationships. The right side illustrates the distinct weight distributions for the first keypoints in different categories.}
\label{fig:KCR2}
\end{figure}

\vspace{5pt}
\noindent\textbf{Weight visualization of Weight Assigner in Keypoint Attention Refiner.} With multiple node relationships in different aspects generated by multiple Attention Filters, the Weight Assigner in KAR is used to selectively keep and discard the relationships. Since the weights are obtained by the corresponding support keypoint token, as mentioned in Section \textcolor{red}{3.4}, therefore support keypoint tokens with similar meaning (left eye of different animal) should yield more similar weights. We visualize the weights in Fig.~\ref{fig:KCR1}, one can see that the similar support keypoint tokens generate similar weights for the $4$ Attention Filters. Likewise, as shown in Fig.~\ref{fig:KCR2}, dissimilar support keypoints bring distinct weight distributions.

\vspace{5pt}
\noindent\textbf{Ablation study on unshared q and k for support keypoint tokens and query images.}
The linear projection before self-attention can be formulated as: 
\begin{equation}\small
    \label{eq:bbox}
    \begin{aligned}
     {K_s}= W_{K1}F_s\,,{K_q}= W_{K2}F_q\,,{K}= \textrm{concat}(K_s,K_q)\,,\\
     {Q_s}= W_{Q1}F_s\,,{Q_q}= W_{Q2}F_q\,,{Q}= \textrm{concat}(Q_s,Q_q)\,,
     \end{aligned}
\end{equation}
where $W_{K1}$, $W_{K2}$, $W_{Q1}$ and $W_{Q2}$ represent the linear projection matrix to generate key $K_s$, $K_q$ and query $Q_s$, $Q_q$ for support feature $F_s$ and query feature $F_q$.
To eliminate the potential impact of increased parameters in Eq.~\eqref{eq:bbox} compared to the vanilla self-attention,  
we conduct an experiment where the linear layers are shared, \ie, $W_{K1}=W_{K2}$ and $W_{Q1}=W_{Q2}$ to maintain the same number of parameters with the vanilla self-attention.
Table \ref{tab:s5} shows that the our method performs comparably either with or without additional parameters, which implies that it is the mechanism rather than the more parameters that boosts the performance. 
\begin{table}\small
\centering
\caption{Ablation study on parameter increase resulting from
Non-Sharing of q and k in Feature Interactor}
\renewcommand\arraystretch{1}
\addtolength{\tabcolsep}{1pt}
 {
\begin{tabular}{@{}lccc@{}}
 \hline
 share&layer&PCK 
\\ \hline
 $\checkmark$&1 &89.1 \\
$\checkmark$ &2&  89.3  \\
\rowcolor{blue!5} & 1 &89.8 \\
\hline
 \end{tabular}
}
\label{tab:s5}
\end{table}

\subsection{Further Exploration for CAPE}

\noindent\textbf{The generality of GKP and KAR.} i) We apply our GKP to the state-of-the-art CapeFormer on split $1$, enhancing PCK by 0.6. ii) By incorporating KAR into the first three layers of the Encoder for interaction, PCK is further improved by 0.8.  iii) The combined effect of these two modules results in an overall enhancement of 1.1 PCK.

\begin{figure}[!t]
\centering{\includegraphics[width=0.9\linewidth]{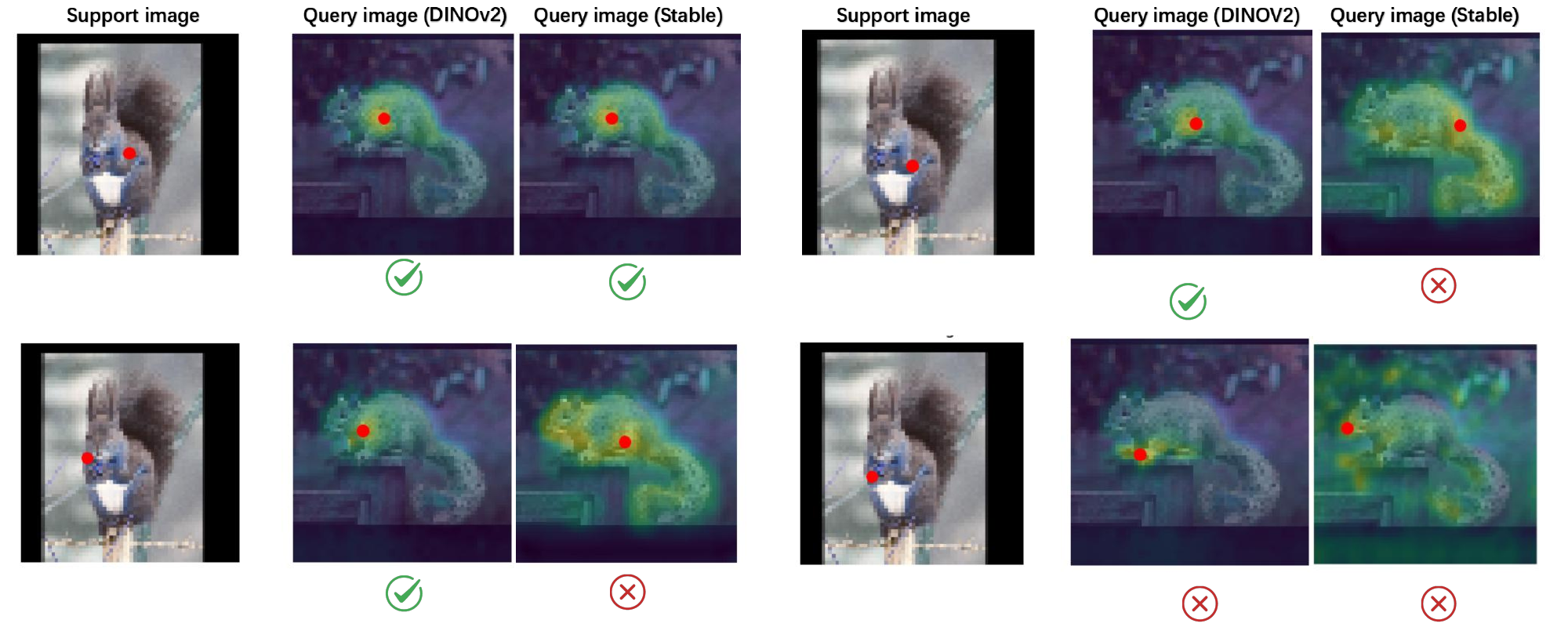}}
\caption{\textbf{Visualization of similarity maps.} The second column corresponds to DINOv2, and the third column represents Stable Diffusion. Red points indicate the final predicted points, with a checkmark denoting correct predictions and a cross indicating incorrect predictions.}
\label{fig:st}
\end{figure}

\noindent\textbf{Whether CAPE-specific models are needed}  

i) Models like VIT, trained in the DINOv2~\cite{oquab2023dinov2} manner, demonstrate fine-grained matching effects across various tasks~\cite{ouyang2022self,stein2024exposing}. Stable Diffusion~\cite{rombach2022high} has also exhibited similar capabilities in DIFT~\cite{tang2023emergent}. We evaluated these approachs on the MP100 dataset following the procedure outlined in DIFT~\cite{tang2023emergent}.  Provided support image and heatmap ground truth for target points, we utilized the visual extractor from Stable Diffusion to extract support features. These features were then multiplied by the heatmap ground truth to obtain support  keypoints. Calculating the cosine similarity between keypoints and the query image produced a similarity map, $S$. We identified the target point P as the $\tt argmax$(S), yielding the final target point. We applied a similar evaluation to DINOv2. We visualized some of these similarity maps, as shown in Fig.~\ref{fig:st}. While they effectively handle simple cases, they face challenges when encountering occlusion or significant appearance differences between support and query images.

\begin{figure}[!t]
\centering{\includegraphics[width=0.7\linewidth]{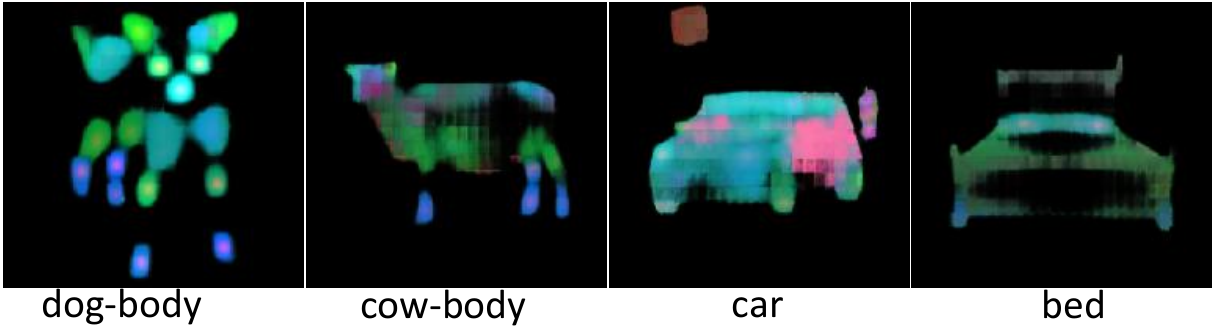}}
\caption{\textbf{Visualization of the rendered style maps for keypoint by Painter.}}
\label{fig:st2}
\end{figure}

ii) 
Several recent papers on generalist models learning assert the ability to solve diverse tasks with just a few task examples, as seen in Painter~\cite{wang2023images} or general approach for visual prompting~\cite{bar2022visual}. It would be interesting to evaluate whether CAPE-specific models are even needed. Here we show the result of generalist model (‘Images Speak in Images’)~\cite{wang2023images} on MP-100 with identical metrics and settings. As shown in the table below, the performance is significantly lower than the CAPE methods. Per visualizations in Fig.~\ref{fig:st2}, the model can only output the shape of furniture and vehicles, but fails to precisely localize the points. Hence, we believe that generalist model cannot replace CAPE models at this moment.  The evaluation code is available in the supplementary material folder.

The results of the three approaches are reported in  Table ~\ref{tab:sla} using consistent metrics. Through a combination of quantitative metrics and visualizations, it becomes evident that the proprietary model of CAPE is needed. Moreover, these general models are notably heavier compared to SCAPE.

\begin{table}[!t]\small
\centering
\caption{\textbf{Results for various general models on MP100 split1.} Specifically, Stable Diffusion performs better when provided with textual cues (right eye) compared to cases without prompts.}
\renewcommand\arraystretch{1}
\addtolength{\tabcolsep}{10pt}
 {
\begin{tabular}{@{}lcc@{}}
 \hline
 Method&spilt1
\\ \hline
DINOv2-L &78.1 \\
DIFT &60.0 \\
DIFT (propmt) &  66.8  \\
Painter &23.9 \\
SCAPE (ours) &91.6 \\
\hline
 \end{tabular}
}
\label{tab:sla}
\end{table}

\subsection{The applicability of SCAPE on more scenarios}
We tested the applicability of our method in cross-category multiple-instance scenarios as well as cross-style scenarios. Indeed, in an effort to seek the most similar keypoint 
of support keypoint, the \textbf{current CAPE} only finds one-to-one correspondence and cannot establish one-to-many correspondence. As shown below, we have tested the applicability of our approach on 
multiple-instance scenarios across categories (Row 1-3) and 
even across domains (Row 4). We think the \textbf{future CAPE} could 
certainly benefit from datasets and frameworks of multi-instance scenarios (including one-to-many correspondence of the same categories). 

\begin{figure*}[!h]
\centering{\includegraphics[width=0.95\linewidth]{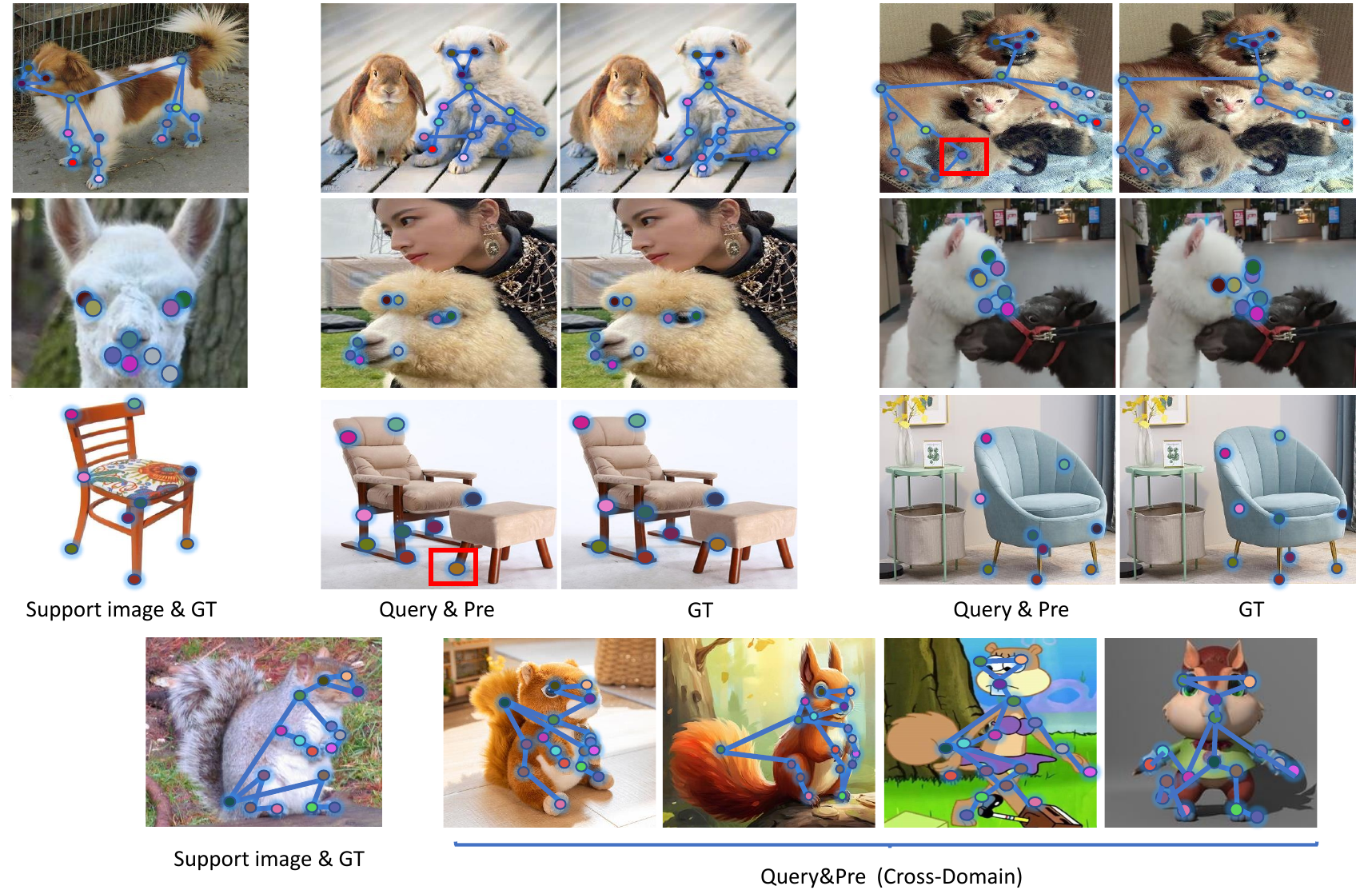}}
\label{fig:com}
\caption{\textbf{  The applicability of SCAPE on multiple-instance scenarios across categories}}
\end{figure*}

\subsection{The code  for SCAPE and testing for Painter} The code of SCAPE and testing for Painter can be found in the attachment.

\subsection{More Qualitative Results}

\begin{figure*}
\centering{\includegraphics[width=1\textwidth,height=1.25\textwidth]{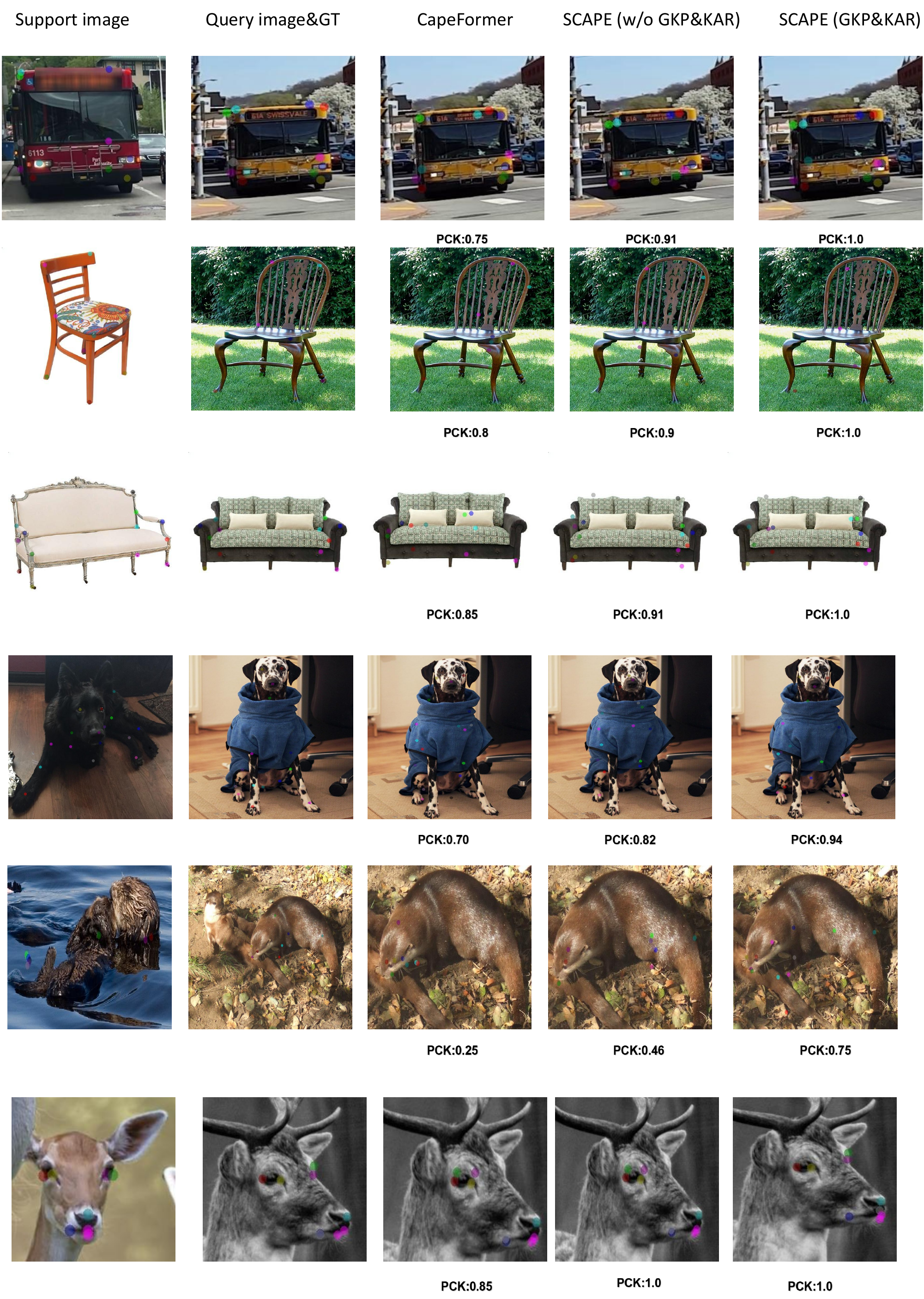}}
\caption{\textbf{ Qualitative results. \#1}}
\label{fig:V1}
\end{figure*}

\begin{figure*}
\centering{\includegraphics[width=1.0\linewidth]{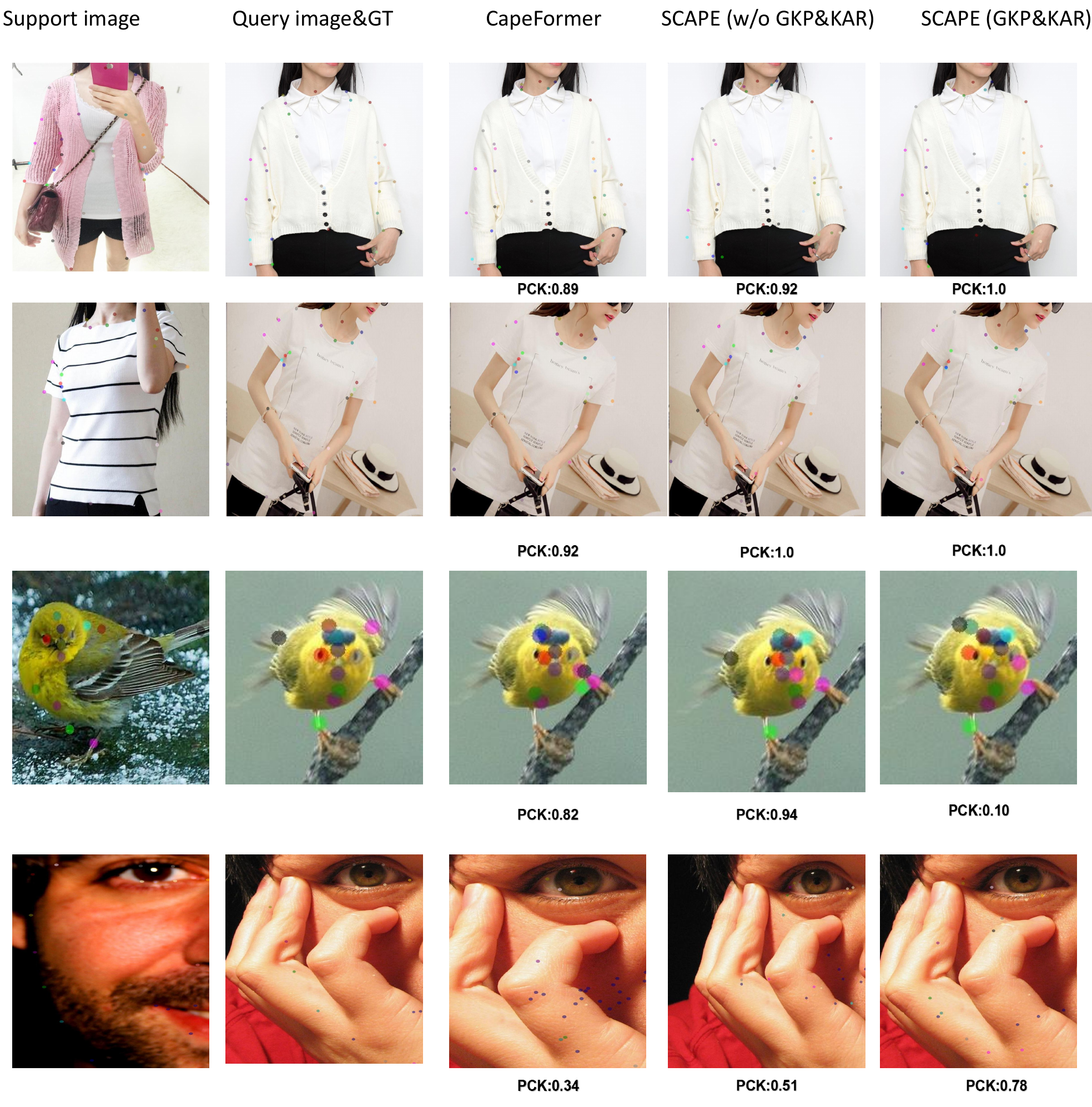}}
\caption{\textbf{ Qualitative results. \#2}}
\label{fig:V2}
\end{figure*}

More qualitative results for visual comparison between the previous best method CapeFormer and ours are shown in Fig.~\ref{fig:V1} and Fig.~\ref{fig:V2}. 

\subsection{Visualization of Attention Maps}

\begin{figure*}
\centering\includegraphics[width=0.9\textwidth,height=1.2\textwidth]{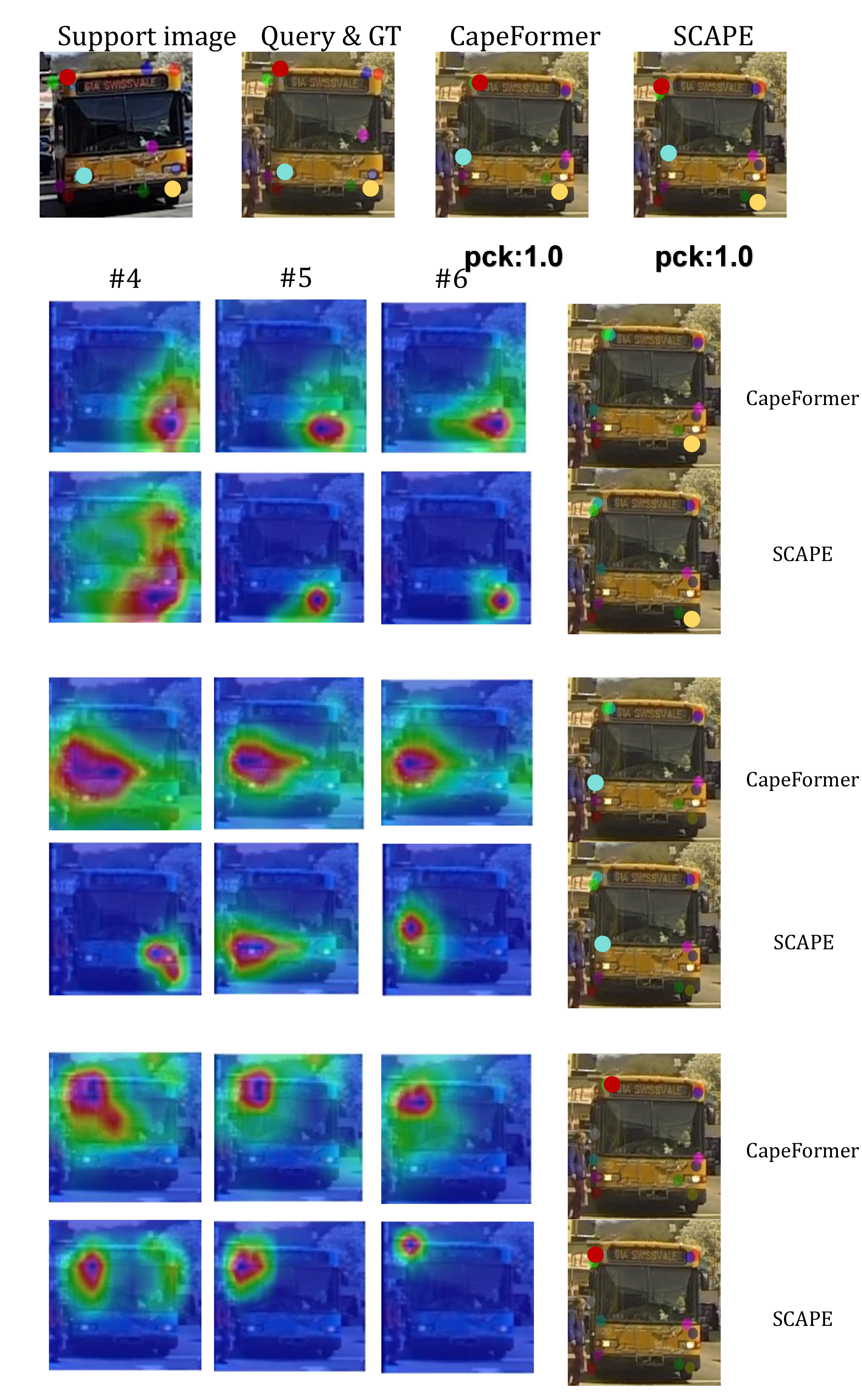}
\caption{\textbf{ Attention map for the last three layers of support keypoints and query image \#1.} In cases where predictions are generally accurate, our method exhibits more convergent attention}
\label{fig:V3}
\end{figure*}

\begin{figure*}
\centering\includegraphics[width=0.9\textwidth,height=1.2\textwidth]{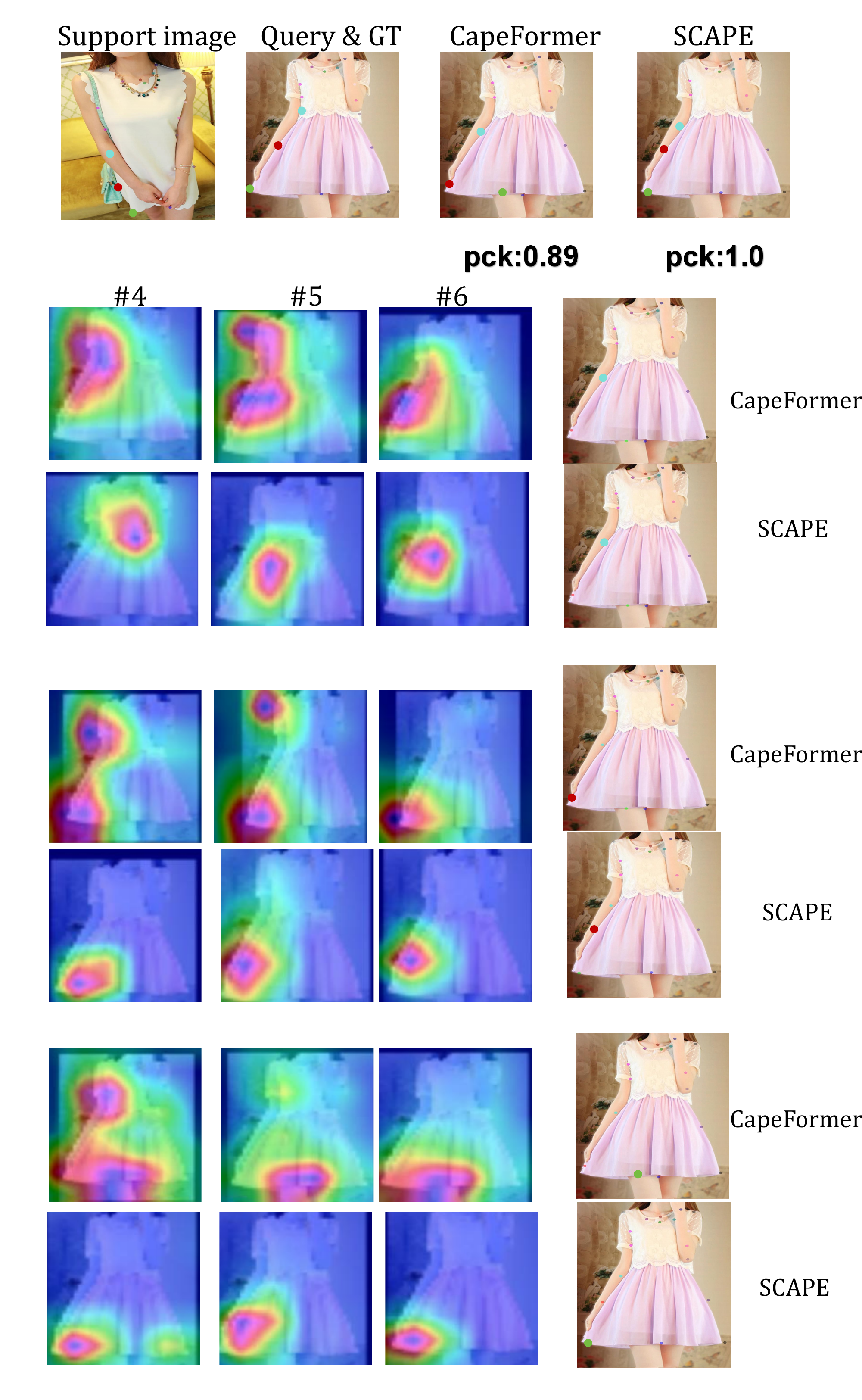}
\caption{\textbf{ Attention map for the last three layers of support keypoints and query image \#2}. SCAPE+ can provide more accurate predictions.}
\label{fig:V4}
\end{figure*}

Fig.~\ref{fig:V3}, Fig.~\ref{fig:V4} depict the attention maps between support keypoints and the query image of the last three layers. The attention maps of both CapeFormer and ours are shown here for better understanding. 

\bibliographystyle{splncs04}
\bibliography{main}
\end{document}